\documentstyle[jair,twoside,11pt,theapa,amsmath,amssymb,amsbsy,amsthm]{article}
\input epsf  %% at top of file
\newcommand{\norm}[1]{||#1||}

\newcommand{\abs}[1]{\lvert#1\rvert}
\newcommand{\set}[1]{\{#1\}}

\newcommand{\jcal}[1]{{\mathcal{#1}}}
\newcommand{\Real}{{\mathbb R}}
\newcommand{\Nat}{{\mathbb N}}
\newcommand{\inner}[2]{\langle#1,#2\rangle}
\DeclareMathOperator{\lspan}{span}
\newcommand{\myalpha}{u}
\newcommand{\vphi}{\phi}
\newcommand{\tphi}{\varphi}
\newcommand{\ttphi}{\tilde{\varphi}}

\newcommand{\func}{\tau}
\newcommand{\basis}{\tilde{\psi}}

\newcommand{\jdim}{n}

\newtheorem{thm}{Theorem}
\newtheorem{lem}{Lemma}
\newtheorem{prp}{Proposition}

\DeclareMathOperator{\argmin}{arg\,min}

\jairheading{}{}{}{}{} \ShortHeadings{On Kernelizing Mahalanobis
Distance Learning Algorithms} {Chatpatanasiri, Korsrilabutr,
Tangchanachaianan \& Kijsirikul} \firstpageno{1}

\begin{document}

\title{On Kernelizing Mahalanobis Distance Learning Algorithms}

\author{\name Ratthachat Chatpatanasiri \email ratthachat.c@student.chula.ac.th \\
\name Teesid Korsrilabutr \email g48tkr@cp.eng.chula.ac.th \\
\name Pasakorn Tangchanachaianan \email pasakorn.t@student.chula.ac.th \\
       \name Boonserm Kijsirikul \email boonserm.k@chula.ac.th \\
       \addr Department of Computer Engineering, Chulalongkorn
       University, Pathumwan, Bangkok, Thailand.}

\maketitle

\begin{abstract}
This paper focuses on the problem of kernelizing an existing
supervised Mahalanobis distance learner. The following features are
included in the paper. Firstly, three popular learners, namely,
``neighborhood component analysis'', ``large margin nearest
neighbors'' and ``discriminant neighborhood embedding'', which do
not have kernel versions are kernelized in order to improve their
classification performances. Secondly, an alternative kernelization
framework called ``KPCA trick'' is presented. Implementing a learner
in the new framework gains several advantages over the standard
framework, e.g. no mathematical formulas and no reprogramming are
required for a kernel implementation, the framework avoids
troublesome problems such as singularity, etc. Thirdly, while the
truths of representer theorems are just assumptions in previous
papers related to ours, here, representer theorems are formally
proven. The proofs validate both the kernel trick and the KPCA trick
in the context of Mahalanobis distance learning. Fourthly, unlike
previous works which always apply brute force methods to select a
kernel, we investigate two approaches which can be efficiently
adopted to construct an appropriate kernel for a given dataset.
Finally, numerical results on various real-world datasets are
presented.
\end{abstract}

%\textbf{Keywords}: Mahalanobis distance, KPCA trick, representer
%theorem, kernel selection.

% DCA, Xing, Yangx2, LPP

\section{Introduction}
\label{sect_intro} Recently, many Mahalanobis distance learners are
invented
\shortcite{Chen:CVPR05,Goldberger:NIPS05,Weinberger:NIPS06,Yang:AAAI06,Sugiyama:ICML06,Yan:PAMI07,Wei:ICML07,Torresani:NIPS07,Xing:NIPS03}.
These recently proposed learners are carefully designed so that they
can handle a class of problems where data of one class form
multi-modality where classical learners such as principal component
analysis (PCA) and Fisher discriminant analysis (FDA) cannot handle.
Therefore, promisingly, the new learners usually outperform the
classical learners on experiments reported in recent papers.
Nevertheless, since learning a Mahalanobis distance is equivalent to
learning a linear map, the inability to learn a non-linear
transformation is one important limitation of all Mahalanobis
distance learners.

As the research in Mahalanobis distance learning has just recently
begun, several issues are left open such as (1) some efficient
learners do not have non-linear extensions, (2) the \emph{kernel
trick} \cite{Scholkopf:BOOK01}, a standard non-linearization method,
is not fully automatic in the sense that new mathematical formulas
have to be derived and new programming codes have to be implemented;
this is not convenient to non-experts, (3) existing algorithms
``assume'' the truth of the \emph{representer theorem}
\shortcite[Chapter 4]{Scholkopf:BOOK01}; however, to our knowledge,
there is no formal proof of the theorem in the context of
Mahalanobis distance learning, and (4) the problem of how to select
an efficient kernel function has been left untouched in previous
works; currently, the best kernel is achieved via a brute-force
method such as cross validation.

In this paper, we highlight the following key contributions:

$\bullet \ $ Three popular learners recently proposed in the
literatures, namely, \emph{neighborhood component analysis} (NCA)
\shortcite{Goldberger:NIPS05}, \emph{large margin nearest neighbors}
(LMNN) \shortcite{Weinberger:NIPS06} and \emph{discriminant
neighborhood embedding} (DNE) \shortcite{Wei:ICML07} are kernelized
in order to improve their classification performances with respect
to the kNN algorithm.

$\bullet \ $ A \emph{KPCA trick} framework is presented as an
alternative choice to the kernel-trick framework. In contrast to the
kernel trick, the KPCA trick does not require users to derive new
mathematical formulas. Also, whenever an implementation of an
original learner is available, users are not required to
re-implement the kernel version of the original learner. Moreover,
the new framework avoids problems such as singularity in
eigen-decomposition and provides a convenient way to speed up a
learner.

$\bullet \ $ Two representer theorems in the context of Mahalanobis
distance learning are proven. Our theorems justify both the
kernel-trick and the KPCA-trick frameworks. Moreover, the theorems
validate kernelized algorithms learning a Mahalanobis distance in
any separable Hilbert space and also cover kernelized algorithms
performing dimensionality reduction.

$\bullet \ $ The problem of efficient kernel selection is dealt
with. Firstly, we investigate the \emph{kernel alignment} method
proposed in previous works \shortcite{Lanckriet:JMLR04,Zhu:NIPS05}
to see whether it is appropriate for a kernelized Mahalanobis
distance learner or not. Secondly, we investigate a simple method
which constructs an unweighted combination of base kernels. A
theoretical result is provided to support this simple approach.
Kernel constructions based on our two approaches require much
shorter running time when comparing to the standard cross validation
approach.

$\bullet \ $ As kNN is already a non-linear classifier, there are
some doubts about the usefulness of kernelizing Mahalanobis distance
learners \shortcite[pp. 8]{Weinberger:NIPS06}. We provide an
explanation and conduct extensive experiments on real-world datasets
to prove the usefulness of the kernelization.

\section{Background}
\label{sect_back} Let $\{\textbf{x}_i, y_i\}_{i=1}^n$ denote a
training set of $n$ labeled examples with inputs $\textbf{x}_i \in
{\mathbb{R}}^D$ and corresponding class labels $y_i \in \{c_1, ...,
c_p\}$. Any Mahalanobis distance can be represented by a symmetric
positive semi-definite (PSD) matrix $M \in {\mathbb{S}}_+^D$. Here,
we denote ${\mathbb{S}}_+^{D}$ as a space of $D \times D$ PSD
matrices. Given two points $\textbf{x}_i$ and $\textbf{x}_j$, and a
PSD matrix $M$, the Mahalanobis distance with respect to $M$ between
the two points is defined as $||\textbf{x}_i - \textbf{x}_j||_M =
\sqrt{(\textbf{x}_i - \textbf{x}_j)^T M (\textbf{x}_i -
\textbf{x}_j)}$. Our goal is to find a PSD matrix $M^*$ that
minimizes a reasonable objective function $f(\cdot)$:
\begin{equation} \label{eq0}
M^* = \underset{M \in {\mathbb{S}}^D_+}{\argmin} f(M).
\end{equation}
Since the PSD matrix $M$ can be decomposed to $A^T A$, we can
equivalently restate our problem as learning the best matrix $A$:
%which minimizes a pre-defined objective function:
\begin{equation} \label{eq1}
A^* = \underset{A \in {\mathbb{R}}^{d \times D}}{\argmin} f(A).
\end{equation}
Note that $d = D$ in the standard setting, but for the purpose of
dimensionality reduction we can learn a low-rank projection by
restricting $d < D$. After learning the best linear map $A^*$, it
will be used by kNN to compute the distance between two points in
the transformed space as $(\textbf{x}_i - \textbf{x}_j)^T M^*
(\textbf{x}_i - \textbf{x}_j)
%\\ = (\textbf{x}_i - \textbf{x}_j)^T A^T A (\textbf{x}_i - \textbf{x}_j)
= || A^*\textbf{x}_i - A^*\textbf{x}_j ||^2$.

In the following subsections, three popular algorithms, whose
objective functions are mainly designed for a further use of kNN
classification, are presented. Despite their efficiency and
popularity, the three algorithms do not have their kernel versions,
and thus in this paper we are primarily interested in kernelizing
these three algorithms in order to improve their classification
performances.

\subsection{Neighborhood Component Analysis (NCA) Algorithm}
\label{sect_nca} The original goal of NCA
\shortcite{Goldberger:NIPS05} is to optimize the
\emph{leave-one-out} (LOO) performance on training data. However, as
the actual LOO classification error of kNN is a non-smooth function
of the matrix $A$, Goldberger et al. propose to minimize a
stochastic variant of the LOO kNN score which is defined as follows:
\begin{equation} \label{eq2}
f^{NCA}(A) = -\sum_i \sum_{y_j = c_i } p_{ij},
\end{equation}
where
\begin{equation*}
p_{ij} = \frac{\exp(- || A\textbf{x}_i - A\textbf{x}_j ||^2)}
{\sum_{k \ne i} \exp(- || A\textbf{x}_i - A\textbf{x}_k ||^2)},\ \ \
\ \ \ \ p_{ii} = 0.
\end{equation*}
Optimizing $f^{NCA}(\cdot)$ can be done by applying a gradient based
method. One major disadvantage of NCA, however, is that
$f^{NCA}(\cdot)$ is not convex, and the gradient based methods are
thus prone to local optima.

\subsection{Large Margin Nearest Neighbor (LMNN) Algorithm}
\label{sect_lmnn} In LMNN \shortcite{Weinberger:NIPS06}, the output
Mahalanobis distance is optimized with the goal that \emph{for each
point, its k-nearest neighbors always belong to the same class while
examples from different classes are separated by a large margin}.

For each point $\textbf{x}_i$, we define its $k$ \emph{target
neighbors} as the $k$ other inputs with the same label $y_i$ that
are closest to $\textbf{x}_i$ (with respect to the Euclidean
distance in the input space). We use $w_{ij} \in \{0,1\}$ to
indicate whether an input $\textbf{x}_j$ is a target neighbor of an
input $\textbf{x}_i$. For convenience, we define $y_{ij} \in
\{0,1\}$ to indicate whether or not the labels $y_i$ and $y_j$
match. The objective function of LMNN is as follows:
\begin{align*}
    f^{LMNN}(M) = \sum_{i,j}w_{ij}||\textbf{x}_i - \textbf{x}_j||_M^2 +
    c\sum_{i,j,l} w_{ij}(1-y_{il})
    \left[1+||\textbf{x}_i - \textbf{x}_j ||_M^2 - ||\textbf{x}_i -
    \textbf{x}_l||_M^2 \right]_+,
\end{align*}
where $[\cdot]_+$ denotes the standard hinge loss: $[z]_+ =
max(z,0)$. The term $c > 0$ is a positive constant typically set by
cross validation. The objective function above is
convex\footnote{There is a variation on LMNN called ``large margin
component analysis'' (LMCA) \cite{Torresani:NIPS07} which proposes
to optimize $A$ instead of $M$; however, LMCA does not preserve some
desirable properties, such as convexity, of LMNN, and therefore the
algorithm ``Kernel LMCA'' presented there is different from ``Kernel
LMNN'' presented in this paper.} and has two competing terms. The
first term penalizes large distances between each input and its
target neighbors, while the second term penalizes small distances
between each input and all other inputs that do not share the same
label.

%The objective function above can be reformulated as an instance of
%\emph{semidefinite programs} (SDPs) \shortcite{Boyd:BOOK04} as shown
%in Figure~\ref{fig_LMNN}. Since SDP is an instance of convex
%programs, in contrast to $f^{NCA}(\cdot)$, the global optimum of
%$f^{LMNN}(\cdot)$ can be efficiently computed. A low-rank
%transformation $A \in \Real^{d \times D}$ such that $d < D$ can be
%achieved by applying the orthogonal decomposition of $M$ and
%retaining only the first $d$ eigenvectors corresponding to the
%smallest $d$ eigenvalues.
%
%\begin{figure}[h]
%\begin{center}
%\vskip -0.20in \fbox{
%\begin{minipage}{10cm}
%  \textbf{Minimize} $\sum_{i,j} w_{ij}(\textbf{x}_i - \textbf{x}_j)^T
%    M (\textbf{x}_i - \textbf{x}_l) + c \sum_{i,j,l} w_{ij} (1-y_{il}) \xi_{ijl}$\\
%\textbf{Subject to:}\\
%    (1) $(\textbf{x}_i - \textbf{x}_l)^T M (\textbf{x}_i -
%    \textbf{x}_l)- (\textbf{x}_i - \textbf{x}_j)^T M (\textbf{x}_i -
%    \textbf{x}_j) \ge 1 - \xi_{ijl}$.\\
%(2) $\xi_{ijl} \ge 0$.\\
%(3) $M \in {\mathbb{S}}^D_+$. \\
%\end{minipage}
%} \vskip -0.1in \caption{an SDP formulation for the LMNN algorithm.}
%\label{fig_LMNN}
%\end{center}
%\vskip -0.3in
%\end{figure}

\subsection{Discriminant Neighborhood Embedding (DNE) Algorithm}
\label{sect_dne} The main idea of DNE \shortcite{Wei:ICML07} is
quite similar to LMNN. DNE seeks a linear transformation such that
neighborhood points in the same class are squeezed but those in
different classes are separated as much as possible. However, DNE
does not care about the notion of margin; in the case of LMNN, we
want every point to stay far from points of other classes, but for
DNE, we want the average distance between two neighborhood points of
different classes to be large. Another difference is that LMNN can
learn a full Mahalanobis distance, i.e. a general \emph{weighted}
linear projection, while DNE can learn only an \emph{unweighted}
linear projection.

Similar to LMNN, we define \emph{two} sets of $k$ target neighbors
for each point $\textbf{x}_i$ based on the Euclidean distance in the
input space. For each $\textbf{x}_i$, let $Neig^I(i)$ be the set of
$k$ nearest neighbors having the same label $y_i$, and let
$Neig^E(i)$ be the set of $k$ nearest neighbors having different
labels from $y_i$. We define $w_{ij}$ as follows
\begin{equation*}
w_{ij} =
\begin{cases}
  +1, & \mbox{if } j \in Neig^I(i) \vee i \in Neig^I(j), \\
  -1, & \mbox{if } j \in Neig^E(i) \vee i \in Neig^E(j), \\
   0, & \mbox{otherwise}.
\end{cases}
\end{equation*}
The objective function of DNE is:
\begin{equation*}
f^{DNE}(A) = \sum_{i,j} w_{i,j} ||A\textbf{x}_i - A\textbf{x}_j||^2.
\end{equation*}
which can be reformulated (up to a constant factor) to be
\begin{equation*}
f^{DNE}(A) = \mbox{trace} (A X (D-W) X^T A^T),
\end{equation*}
where $W$ is a symmetric matrix with elements $w_{ij}$, $D$ is a
diagonal matrix with $D_{ii} = \sum_{j} w_{ij}$ and $X$ is the
matrix of input points $(\textbf{x}_1, ..., \textbf{x}_n)$. It is a
well-known result from spectral graph theory
\cite{vonLuxburg:STATCOMP07} that $D-W$ is symmetric but is not
necessarily PSD. To solve the problem by eigen-decomposition, the
constraint $AA^T = I$ is added (recall that $A \in {\mathbb{R}}^{d
\times D}$ $(d \le D)$) so that we have the following optimization
problem:
\begin{equation} \label{eq_dne}
  A^* = \underset{A A^T = I}{\operatorname{\argmin}}\, \mbox{trace}(AX(D-W)X^TA^T).
\end{equation}
%For the purpose of finding a low-rank projection, recall that $A \in
%{\mathbb{R}}^{d \times D}$ $(d \le D)$. From now on, we will work
%with $A^T \in {\mathbb{R}}^{D \times d}$ instead of $A$ for
%notational convenience. Let $A^T = (\textbf{a}_1, ...,
%\textbf{a}_d)$ where $\textbf{a}_i \in {\mathbb{R}}^D$. By adding
%further constraints that,
%\begin{equation} \label{eq_dne1}
%\textbf{a}_i^T \textbf{a}_i = 1 \  \mbox{ and } \  \textbf{a}_i^T
%\textbf{a}_j = 0 \ (i \ne j),
%\end{equation}
%the optimal vectors $\textbf{a}_1, ..., \textbf{a}_d$ can be found
%by solving the following eigenvalue problem \shortcite[Chapter
%10]{Fukunaga:Book90}:
%\begin{equation*}
%X (D-W) X^T \textbf{a}_i = \lambda_i \textbf{a}_i.
%\end{equation*}
%By rearranging the eigenvalues such that $\lambda_1 \le ... \le
%\lambda_D$, the optimal matrix $(\textbf{a}_1, ..., \textbf{a}_d)$
%can be constructed from the unit eigenvectors corresponding to the
%first $d$ eigenvalues. As LMNN, the global optimum of
%$f^{DNE}(\cdot)$ with constraints specified by Eq.~(\ref{eq_dne1})
%can be efficiently computed.
%
%One advantage of DNE over LMNN and NCA is that, for DNE, we have a
%deterministic rule to select the optimal dimensionality $d$ of the
%transformed space: $d$ will be the number of \emph{negative}
%eigenvalues obtained from the above eigenvalue problem
%\shortcite{Wei:ICML07}.

\section{Kernelization} \label{sect_kernel}
In this section, we focus on two kernelization frameworks going to
non-linearize the three algorithms presented in the previous
section. First, the standard kernel trick framework is presented.
Next, the \emph{KPCA trick} framework which is an alternative to the
kernel-trick framework is presented. Kernelization in this new
framework is conveniently done with an application of \emph{kernel
principal component analysis} (KPCA). Finally, representer theorems
are proven to validate all applications of the two kernelization
frameworks in the context of Mahalanobis distance learning. Note
that in some previous works
\shortcite{Chen:CVPR05,Globerson:NIPS06,Yan:PAMI07,Torresani:NIPS07},
the validity of applications of the the kernel trick has not been
proven.

\subsection{Historical Background}
After finishing writing the current paper, we just knew that this
name was first appeared in the paper of \citeA{Chapelle:NIPS01} who
first applied this method to invariant support vector machines;
moreover, it is appeared to us that the KPCA trick has been known to
some researchers (private communication to some ECML reviewers).
Without knowing about this fact, we reinvented the framework and,
coincidentally, called it ``KPCA trick'' ourselves. Nevertheless, we
will shown in Section~\ref{sect_versus} that, in the context of
Mahalanobis distance learning, the KPCA trick non-trivially has many
advantages over the kernel trick; we believe that this consequence
is new and is not a consequence of previous works. Also,
mathematical tools provided in previous works
\shortcite{Scholkopf:BOOK01,Chapelle:NIPS01} are not enough to prove
the validity of the KPCA trick in this context, and thus the new
validation proof of the KPCA trick is needed (see our
Theorem~\ref{thm_repr}).

\subsection{The Kernel-Trick Framework} \label{sect_framework}
Given a PSD kernel function $k(\cdot,\cdot)$
\shortcite{Scholkopf:BOOK01}, we denote $\phi$, $\phi'$ and $\phi_i$
as mapped data (in a feature space associated with the kernel) of
each example $\textbf{x}$, $\textbf{x}'$ and $\textbf{x}_i$,
respectively. A (squared) Mahalanobis distance under a matrix $M$ in
the feature space is
\begin{align} \label{eq_framework1}
(\phi_i - \phi_j)^T M (\phi_i - \phi_j) =
  (\phi_i - \phi_j)^T A^T A (\phi_i - \phi_j).
\end{align}
To be consistent with Subsection~\ref{sect_dne}, let $A^T =
(\textbf{a}_1, ..., \textbf{a}_d)$. Denote a (possibly
infinite-dimensional) matrix of the mapped training data $\Phi =
(\phi_1, ..., \phi_n)$. The main idea of the kernel-trick framework
is to parameterize (see representer theorems below)
\begin{equation} \label{eq_framework2}
A^T = \Phi U^T,
\end{equation}
where $U^T = (\textbf{u}_1, ..., \textbf{u}_d)$. Substituting $A$ in
Eq.~(\ref{eq_framework1}) by using Eq.~(\ref{eq_framework2}), we
have
\begin{align*}
(\phi_i - \phi_j)^T M (\phi_i - \phi_j) = (\textbf{k}_i -
\textbf{k}_j)^T U^T U (\textbf{k}_i - \textbf{k}_j),
\end{align*}
where
\begin{equation} \label{eq_framework3}
\textbf{k}_i = \Phi^T \phi_i = \big(\langle \phi_1, \phi_i\rangle,
..., \langle\phi_n, \phi_i \rangle \big)^T.
\end{equation}
Now our formula depends only on an inner-product $\langle\phi_i,
\phi_j\rangle$, and thus the kernel trick can be now applied by
using the fact that $k(\textbf{x}_i, \textbf{x}_j) = \langle\phi_i,
\phi_j\rangle$ for a PSD kernel function $k(\cdot,\cdot)$.
Therefore, the problem of finding the best Mahalanobis distance in
the feature space is now reduced to finding the best linear
transformation $U$ of size $d \times n$. Nonetheless, it often
happens that finding $U$ is much more troublesome than finding $A$
in the input space, even their optimization problems look similar,
as shown in Section~\ref{sect_versus}.

Once we find the matrix $U$, the Mahalanobis distance from a new
test point $\textbf{x}'$ to any input point $\textbf{x}_i$ in the
feature space can be calculated as follows:
\begin{equation} \label{eq_framework4}
|| \phi' - \phi_i||^2_M = (\textbf{k}' - \textbf{k}_i)^T U^T U
(\textbf{k}' - \textbf{k}_i),
\end{equation}
where $\textbf{k}' = (k(\textbf{x}', \textbf{x}_1), ...,
k(\textbf{x}', \textbf{x}_n))^T$. kNN classification in the feature
space can be performed based on Eq.~\eqref{eq_framework4}.

\subsection{The KPCA-Trick Framework}
\label{sect_kpca} As we emphasize above, although the kernel trick
framework can be applied to non-linearize the three learners
introduced in Section~\ref{sect_back}, it often happens that finding
$U$ is much more troublesome than finding $A$ in the input space,
even their optimization problems look similar (see
Section~\ref{sect_versus}). In this section, we develop a \emph{KPCA
trick} framework which can be much more conveniently applied to
kernelize the three learners.

Denote $k(\cdot,\cdot)$, $\phi_i$, $\phi$ and $\phi'$ as in
Subsection~\ref{sect_framework}. The central idea of the KPCA trick
is to represent each $\phi_i$ and $\phi'$ in a new
``finite''-dimensional space, without any loss of information.
Within the framework, a new coordinate of each example is computed
``explicitly'', and each example in the new coordinate is then used
as the input of any existing Mahalanobis distance learner. As a
result, by using the KPCA trick in place of the kernel trick, there
is no need to derive new mathematical formulas and no need to
implement new algorithms.

To simplify the discussion of KPCA, we assume that $\{\phi_i\}$ is
linearly independent and has its center at the origin, i.e. $\sum_i
\phi_i = 0$ (otherwise, $\{\phi_i\}$ can be centered by a simple
pre-processing step \cite[p. 115]{Shawe:BOOK04}). Since we have $n$
total examples, the span of $\{\phi_i\}$ has dimensionality $\jdim$.
Here we claim that each example $\phi_i$ can be represented as
$\tphi_i \in \Real^\jdim$ with respect to a new \emph{orthonormal}
basis $\{\psi_i\}_{i=1}^\jdim$ such that
$span(\{\psi_i\}_{i=1}^\jdim)$ is the same as
$span(\{\phi_i\}_{i=1}^n)$ without loss of any information. More
precisely, we define
\begin{equation} \label{eq_newcoord}
    \tphi_i = \Big( \inner{\phi_i}{\psi_1}, \hdots, \inner{\phi_i}{\psi_\jdim} \Big) = \Psi^T \phi_i.
\end{equation}
where $\Psi = (\psi_1, ..., \psi_\jdim)$. Note that although we may
be unable to numerically represent each $\psi_i$, an inner-product
of $\inner{\phi_i}{\psi_j}$ can be conveniently computed by KPCA (or
kernel Gram-Schmidt \shortcite{Shawe:BOOK04}). Likewise, a new test
point $\phi'$ can be mapped to $\tphi' = \Psi^T \phi'$.
Consequently, the mapped data $\{\tphi_i\}$ and $\tphi'$ are
finite-dimensional and can be explicitly computed.

\subsubsection{The KPCA-trick Algorithm}
\begin{figure}[h]
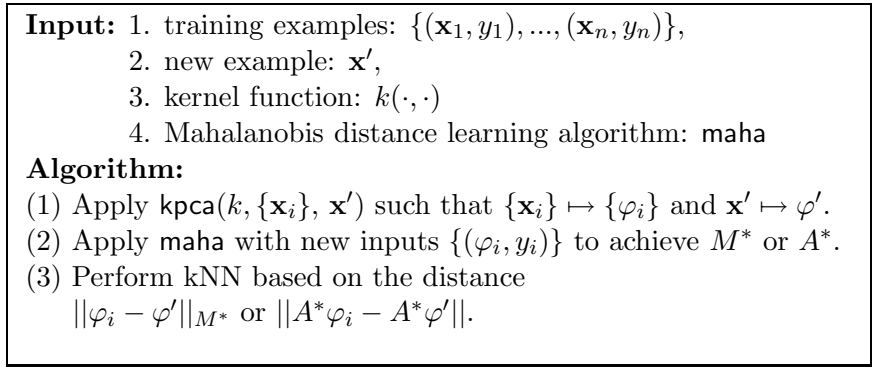

\begin{center}
\vskip -0.2in \fbox{
\begin{minipage}{11cm}
  \textbf{Input:} 1. training examples: $\{(\textbf{x}_1,y_1), ..., (\textbf{x}_n,y_n)\},$\\
  \hspace*{1.25cm} 2. new example: $\textbf{x}'$,\\
  \hspace*{1.25cm} 3. kernel function: $k(\cdot,\cdot)$\\
  \hspace*{1.25cm} 4. Mahalanobis distance learning algorithm: \textsf{maha}\\
\textbf{Algorithm:}\\
    (1) Apply \textsf{kpca}($k,\{\textbf{x}_i\}$, $\textbf{x}'$) such that $\{\textbf{x}_i\} \mapsto \{\tphi_i\}$
    and $\textbf{x}' \mapsto \tphi'$.\\
    (2) Apply \textsf{maha} with new inputs $\{(\tphi_i,y_i)\}$ to achieve $M^*$ or $A^*$.\\
    (3) Perform kNN based on the distance\\ $\hspace*{0.5cm}$ $\norm{\tphi_i - \tphi'}_{M^*}$ or $\norm{A^*\tphi_i - A^*\tphi'}$.\\
\end{minipage}
} \vskip -0.1in \caption{The KPCA-trick algorithm.} \label{fig_KPCA}
\end{center}
\vskip -0.3in
\end{figure}

The KPCA-trick algorithm consisting of three simple steps is shown
in Figure~\ref{fig_KPCA}. NCA, LMNN, DNE and other learners,
including those in other settings (e.g. semi-supervised settings),
whose kernel versions are previously unknown
\shortcite{Yang:AAAI06,Xing:NIPS03,Me:Arxiv08b} can all be
kernelized by this simple algorithm. Therefore, it is much more
convenient to kernelize a learner by applying the KPCA-trick
framework rather than applying the kernel-trick framework. In the
algorithm, we denote a Mahalanobis distance learner by \textsf{maha}
which performs the optimization process shown in Eq.~\ref{eq0} (or
Eq.~\ref{eq1}) and outputs the best Mahalanobis distance $M^*$ (or
the best linear map $A^*$).

\subsubsection{Representer Theorems}
Is it valid to represent an infinite-dimensional vector $\phi$ by a
finite-dimensional vector $\tphi$? In the context of SVMs
\shortcite{Chapelle:NIPS01}, this validity of the KPCA trick is
easily achieved by straightforwardly extending a proof of an
established representer theorem
\shortcite{Scholkopf:COLT01}\footnote{A representer theorem, along
with Mercer theorem, is a key ingredient for validating the kernel
trick \shortcite{Scholkopf:BOOK01}. The origin of the classical
representer theorem is dated back to at least 1970s
\cite{Wahba:jmma71}.}. In the context of Mahalanobis distance
learning, however, proofs provided in previous works cannot be
directly extended. Note that, in the SVM cases considered in
previous works, what is learned is a hyperplane, a linear functional
outputting a 1-dimensional value. In our case, as shown in
Eq.~\ref{eq_framework2}, what is learned is a linear map which, in
general, outputs a \emph{countably infinite dimensional} vector.
Hence, to prove the validity of the KPCA trick in our case, we need
some mathematical tools which can handle a countably infinite
dimensionality. Below we give our versions of representer theorems
which prove the validity of the KPCA trick in the current context.

By our representer theorems, it is the fact that, given an objective
function $f(\cdot)$ (see Eq.~\eqref{eq0}), the optimal value of
$f(\cdot)$ based on the input $\{\phi_i\}$ is equal to the optimal
value of $f(\cdot)$ based on the input $\{\tphi_i\}$. Hence, the
representation of $\tphi_i$ can be safely applied. We separate the
problem of Mahalanobis distance learning into two different cases.
The first theorem covers Mahalanobis distance learners (learning a
full-rank linear transformation) while the second theorem covers
dimensionality reduction algorithms (learning a low-rank linear
transformation).

\begin{thm} (Full-Rank Representer Theorem) \label{thm_repr}
  Let $\set{\basis_i}_{i=1}^\jdim$ be a set of points in a feature
  space $\jcal{X}$ such that $span(\set{\basis_i}_{i=1}^\jdim)$ =
  $span(\set{\phi_i}_{i=1}^n)$, and $\jcal{X}$ and $\jcal{Y}$ be separable Hilbert spaces.
  For an objective function $f$ depending only on $\set{\inner{A\vphi_i}{A\vphi_j}}$, the optimization
  \begin{align*}
    \min_{A}: \ \  &f(\inner{A\vphi_1}{A\vphi_1},\ldots,\inner{A\vphi_i}{A\vphi_j},\ldots,\inner{A\vphi_n}{A\vphi_n}) \\
    \textrm{s.t.} \ & A:\jcal{X}\to\jcal{Y} \textrm{ is a bounded linear map}
    \enspace,
  \end{align*}
  has the same optimal value as,
  \begin{align*}
    \min_{A' \in \Real^{\jdim \times \jdim}}
    f(\ttphi_1^T A'^T A' \ttphi_1,\ldots,\ttphi_i^T A'^T A' \ttphi_j,\ldots,\ttphi_n^T A'^T A' \ttphi_n),
    %\textrm{s.t.}: & G \textrm{ is an } M\times M \textrm{ PSD matrix}
    %\enspace.
  \end{align*}
  where
  $\ttphi_i=\left(\inner{\vphi_i}{\basis_1},\ldots,\inner{\vphi_i}{\basis_\jdim}\right)^T \in \Real^\jdim$.
\end{thm}

To our knowledge, mathematical tools provided in the work of
\shortciteA[Chapter 4]{Scholkopf:BOOK01} are not enough to prove
Theorem~\ref{thm_repr}. The proof presented here is a
non-straightforward extension of
\shortciteauthor{Scholkopf:BOOK01}'s work. We also note that
Theorem~\ref{thm_repr}, as well as Theorem~\ref{thm_repr2} shown
below, is more general than what we discuss above. They justify both
the kernel trick (by substituting $\basis_i = \phi_i$ and hence
$\ttphi_i = \textbf{k}_i$) and the KPCA trick (by substituting
$\basis_i = \psi_i$ and hence $\ttphi_i = \tphi_i$).

To start proving Theorem~\ref{thm_repr}, the following lemma is
useful.
\begin{lem}\label{thm:riesz}
  Let $\jcal{X},\jcal{Y}$ be two Hilbert spaces and $\jcal{Y}$ is separable, i.e.~$\jcal{Y}$ has a
  countable orthonormal basis $\{e_i\}_{i\in\Nat}$. Any bounded linear
  map $A:\jcal{X}\to \jcal{Y}$ can be uniquely decomposed as $\sum_{i=1}^\infty
  \langle\cdot,\func_i\rangle_\jcal{X} e_i$ for some $\{\func_i\}_{i\in\Nat}\subseteq
  \mathcal{X}$.
\end{lem}
\begin{proof}
%  \emph{Proof.}
  As $A$ is bounded, the linear functional
  $\vphi \mapsto \langle {A\vphi,e_i\rangle}_\jcal{Y}$ is bounded for every $i$
  since, by Cauchy-Schwarz inequality,
%  \begin{equation*}
    $\abs{\langle A\vphi,e_i\rangle_\jcal{Y}} \le
    \norm{A\vphi} \norm{e_i} \le
    \norm{A} \norm{\vphi}.$ % \enspace.
%  \end{equation*}
  By Riesz representation theorem, the map $\langle A\cdot,e_i\rangle_\jcal{Y}$ can be
  written as $\langle \cdot,\func_i\rangle_\jcal{X}$ for a unique $\func_i\in \jcal{X}$.
  Since $\{e_i\}_{i\in\Nat}$ is an orthonormal basis of $\jcal{Y}$,
  for every $\vphi\in \jcal{X}$,
  $
    A\vphi = \sum_{i=1}^\infty\langle A\vphi,e_i\rangle_\jcal{Y} e_i
    = \sum_{i=1}^\infty\langle \vphi,\func_i\rangle_\jcal{X} e_i
  $.
\end{proof}

\begin{proof} (Theorem~\ref{thm_repr})
  To avoid complicated notations, we omit subscripts such as
  $\jcal{X},\jcal{Y}$ of inner products.
  The proof will consist of two steps. In the first step, we will
  prove the theorem by assuming that $\set{\basis_i}_{i=1}^\jdim$ is an
  orthonormal set. In the second step, we prove the theorem in
  general cases where $\set{\basis_i}_{i=1}^\jdim$ is not necessarily orthonormal.
  The proof of the first step requires an application of Fubini theorem \shortcite{Wicharn:BOOK06}.
%  \begin{equation*}
%    A\vphi' = \sum_{i=1}^\infty \inner{\vphi'}{\func_i}e_i.
%  \end{equation*}

  \textbf{Step 1}. Assume that $\set{\basis_i}_{i=1}^\jdim$ is an
  orthonormal set.  Let $\{e_i\}_{i=1}^\infty$ be an orthonormal basis of $\jcal{Y}$.
  For any $\vphi'\in \jcal{X}$, we have, by
  Lemma~\ref{thm:riesz}, $A\vphi' = \sum_{k=1}^\infty \inner{\vphi'}{\func_k}e_k$.
  Hence, for each bounded linear map $A:\jcal{X}\to\jcal{Y}$, and $\vphi,\vphi'\in
  span(\set{\basis_i}_{i=1}^\jdim)$, we have
%  \begin{align*}
    $\inner{A\vphi}{A\vphi'} =
    \sum_{k=1}^\infty\inner{\vphi}{\func_k}\inner{\vphi'}{\func_k}.$
%  \end{align*}

  Note that Each $\func_k$ can be
  decomposed as $\func'_k+\func_k^{\bot}$ such that $\func_k'$ lies in
  $span(\set{\basis_i}_{i=1}^\jdim)$ and $\func_k^{\bot}$ is orthogonal to the
  span. These facts make $\inner{\vphi'}{\func_k}=\inner{\vphi'}{\func_k'}$ for every $k$.
  Moreover, $\func'_k = \sum_{j=1}^\jdim\myalpha_{kj}\basis_j$, for some $\{u_{k1}, ...,u_{k\jdim}\} \subset \Real^\jdim$. Hence, we have
  \begin{align*}
    \inner{A\vphi}{A\vphi'} =
    \sum_{k=1}^\infty\inner{\vphi}{\func_k}\inner{\vphi'}{\func_k}
    &= \sum_{k=1}^\infty\inner{\vphi}{\func'_k}\inner{\vphi'}{\func'_k}\\
    &= \sum_{k=1}^\infty\inner{\vphi}{\sum_{i=1}^\jdim\myalpha_{ki}\basis_i}\inner{\vphi'}{\sum_{i=1}^\jdim\myalpha_{ki}\basis_i} \\
    &= \sum_{k=1}^\infty \sum_{i,j=1}^\jdim \myalpha_{ki}\myalpha_{kj}\inner{\vphi}{\basis_i}\inner{\vphi'}{\basis_j} \\
    (\mbox{Fubini theorem: explained below}) &= \sum_{i,j=1}^\jdim \left(\sum_{k=1}^\infty \myalpha_{ki}\myalpha_{kj}\right)\inner{\vphi}{\basis_i}\inner{\vphi'}{\basis_j} \\
    &= \sum_{i,j=1}^\jdim G_{ij}\inner{\vphi}{\basis_i}\inner{\vphi'}{\basis_j}\\
    &= \ttphi^TG \ttphi' = \ttphi^T A'^T A' \ttphi'.
  \end{align*}
  At the fourth equality, we apply Fubini theorem to swap the two
  summations. To see that Fubini theorem can be applied at the fourth equality, we
  first note that $\sum_{k=1}^\infty \myalpha_{ki}^2$ is finite for each $i\in\set{1\ldots \jdim}$ since
  \begin{equation*}
    \sum_{k=1}^\infty\myalpha_{ki}^2 =
    \sum_{k=1}^\infty\inner{\basis_i}{\sum_{j=1}^\jdim\myalpha_{kj}\basis_j}\inner{\basis_i}{\sum_{j=1}^\jdim\myalpha_{kj}\basis_j}=
    \norm{A\basis_i}^2 < \infty.
  \end{equation*}
  Applying the above result together with Cauchy-Schwarz inequality and Fubini theorem for non-negative
  summation, we have
  \begin{align*}
    \sum_{k=1}^\infty \sum_{i,j=1}^\jdim |\myalpha_{ki}\myalpha_{kj}\inner{\vphi}{\basis_i}\inner{\vphi'}{\basis_j}|
    &= \sum_{i,j=1}^\jdim  \sum_{k=1}^\infty |\myalpha_{ki}\myalpha_{kj}\inner{\vphi}{\basis_i}\inner{\vphi'}{\basis_j}|\\
    &= \sum_{i,j=1}^\jdim  |\inner{\vphi}{\basis_i}\inner{\vphi'}{\basis_j}| \Big(\sum_{k=1}^\infty |\myalpha_{ki}\myalpha_{kj}|\Big)\\
    &\le \sum_{i,j=1}^\jdim  |\inner{\vphi}{\basis_i}\inner{\vphi'}{\basis_j}| \sqrt{\Big(\sum_{k=1}^\infty \myalpha_{ki}^2\Big) \Big(\sum_{k=1}^\infty \myalpha_{kj}^2\Big)}\\
    &< \infty.
  \end{align*}
  Hence, the summation converges absolutely and thus Fubini theorem can be applied as claimed above.
  Again, using the fact that $\sum_{k=1}^\infty \myalpha_{ki}^2 <
  \infty$, we have that each element of
  $G$, $G_{ij} = \sum_{k=1}^\infty \myalpha_{ki}\myalpha_{kj}$, is finite.  Furthermore, the matrix $G$ is PSD since
  each of its elements can be regarded as an inner product of two vectors
  in $\ell_2$.

  Hence, we finally have that $\inner{A\vphi_i}{A\vphi_j}=\ttphi_i^T A'^T A' \ttphi_j$,
  for each $1 \le i,j \le n$.
  Hence, whenever a map $A$ is given, we can construct $A'$ such that it results in the
  same objective function value. By reversing the proof, it is easy to see that the
  converse is also true. The first step of the proof is finished.

  \textbf{Step 2}. We now prove the theorem without assuming that $\set{\basis_i}_{i=1}^\jdim$ is an
  orthonormal set. Let all notations be the same as in Step 1. Let $\Psi'$ be the matrix
  $(\basis_1, ..., \basis_\jdim)$. Define $\set{\psi_i}_{i=1}^\jdim$ as
  an orthonormal set such that $span(\set{\psi_i}_{i=1}^\jdim)$ =
  $span(\set{\basis_i}_{i=1}^\jdim)$ and $\Psi = (\psi_1, ..., \psi_\jdim)$
  and $\varphi_i = \Psi^T \phi_i$. Then,
  we have that $\basis_i = \Psi \textbf{c}_i$ for some $\textbf{c}_i \in \Real^\jdim$
  and $\Psi' = \Psi C$ where $C = (\textbf{c}_1, ...,\textbf{c}_\jdim)$.
  Moreover, since $C$ map from an independent set
  $\set{\psi_i}$ to another independent set $\set{\basis_i}$, $C$ is invertible.
  We then have, for any $A'$,
  \begin{align*}
  \ttphi_i^T  A'^T A' \ttphi_j
     &= \phi_i^T \Psi' A'^T A' \Psi'^T \phi_j\\
     &= \phi_i^T \Psi C A'^T A' C^T \Psi^T \phi_j\\
     &= \varphi_i^T C A'^T A' C^T \varphi_j\\
     &= \varphi_i^T B^T B \varphi_j,
  \end{align*}
  where $A' C^T = B$ and $A' = B (C^T)^{-1}$. Hence, for any $B$ we
  have the matrix $A'$ which gives $\ttphi_i^T  A'^T A'
  \ttphi_j = \varphi_i^T B^T B \varphi_j$. Using the same arguments as
  in Step 1, we finish the proof of Step 2 and of Theorem~\ref{thm_repr}.
\end{proof}

\begin{thm} (Low-Rank Representer Theorem) \label{thm_repr2}
  Define $\set{\basis_i}_{i=1}^\jdim$  and $\ttphi_i$ be as in Theorem~\ref{thm_repr}.
  an objective function $f$ depending only on $\set{\inner{A\vphi_i}{A\vphi_j}}$, the optimization
  \begin{align*}
    \min_{A}: \ \  &f(\inner{A\vphi_1}{A\vphi_1},\ldots,\inner{A\vphi_i}{A\vphi_j},\ldots,\inner{A\vphi_n}{A\vphi_n}) \\
    \textrm{s.t.} \ & A:\jcal{X}\to\Real^d \textrm{ is a bounded linear map}
    \enspace,
  \end{align*}
  has the same optimal value as,
  \begin{align*}
    \min_{A' \in \Real^{d \times \jdim}}
    f(\ttphi_1^T A'^T A' \ttphi_1,\ldots,\ttphi_i^T A'^T A' \ttphi_j,\ldots,\ttphi_n^T A'^T A'
    \ttphi_n).
    %\textrm{s.t.}: & G \textrm{ is an } M\times M \textrm{ PSD matrix}
    %\enspace.
  \end{align*}
\end{thm}
The proof of Theorem~\ref{thm_repr2} is a generalization of the
proofs in previous works \cite[Chap. 4]{Scholkopf:BOOK01}.

\begin{proof} (Theorem~\ref{thm_repr2})
  Let $\{e_i\}_{i=1}^d$ be the canonical basis of $\Real^d$, and let
  $\vphi\in\lspan\set{\basis_1,\ldots\,\basis_n}$. By Lemma~\ref{thm:riesz},
  $ A\vphi = \sum_{i=1}^d\inner{\vphi}{\tau_i}e_i$
  for some $\tau_1,\ldots,\tau_d\in \jcal{X}$.  Each $\tau_i$ can be
  decomposed as $\tau'_i+\tau_i^{\bot}$ such that $\tau_i'$ lies in
  $\lspan\set{\basis_1,\ldots,\basis_n}$ and $\tau_i^{\bot}$ is orthogonal to the
  span. These facts make $\inner{\vphi}{\tau_i}=\inner{\vphi}{\tau_i'}$ for every $i$. We
  then have, for some $\myalpha_{ij}\in{\Real}, \ 1 \le i \le d, \ 1 \le j\le n$,
  \begin{align*}
    A\vphi &= \sum_{i=1}^d\inner{\vphi}{\sum_{j=1}^n\myalpha_{ij}\basis_j}e_i
    &= \sum_{i=1}^de_i\sum_{j=1}^n\myalpha_{ij}\inner{\vphi}{\basis_j}
    &=
    \begin{bmatrix}
      \myalpha_{11} & \cdots & \myalpha_{1n} \\
      \vdots & \ddots & \vdots \\
      \myalpha_{d1} & \cdots & \myalpha_{dn}
    \end{bmatrix}
    \begin{bmatrix}
      \inner{\vphi}{\basis_1} \\
      \vdots \\
      \inner{\vphi}{\basis_n}
    \end{bmatrix}
    = U\ttphi
    \enspace.
  \end{align*}
  Since every $\vphi_i$ is in the span, we conclude that $A\vphi_i=U\ttphi_i$.
  Now, one can easily check that $\inner{A\vphi_i}{A\vphi_j}=\ttphi_i^TU^TU\ttphi_j$.
  Hence, whenever a map $A$ is given, we can construct $U$ such that it results in the
  same objective function value. By reversing the proof, it is easy to see that the
  converse is also true, and thus the theorem is proven (by renaming $U$ to $A'$).
\end{proof}
Note that the proof of Theorem~\ref{thm_repr2} cannot be directly
used for proving Theorem~\ref{thm_repr} (let $d = \infty$ and $U \in
\Real^{\infty \times n}$, and Theorem~\ref{thm_repr2} is still
valid. However, to practically be useful, we need a
finite-dimensional linear map. Hence, we must show that $U^T U \in
\Real^{n \times n}$ by proving that $u_{ij} < \infty$ for all
$i,j$).

%\noindent \textbf{Remark}. In Theorem~\ref{thm_repr2}, we can also
%write $A\vphi  = U'\Phi^T\vphi$ for some $U' \in R^{}$. One
%immediate corollary which is already used in eq. (16) is that $A^T$
%can be represented by $\Phi U^T$.
%\newline
%The proof of the theorems will be given in the next section.\\\\
\subsubsection{Remarks} \label{sect_remark}
1. Note that by Mercer theorem \shortcite[pp. 37]{Scholkopf:BOOK01},
we can either think of each $\phi_i \in \ell_2$ or $\phi_i \in
\Real^N$ for some positive integer $N$, and thus the assumption of
Theorem~\ref{thm_repr} that $\jcal{X}$, as well as $\jcal{Y}$, is
separable Hilbert space is then valid. Also, both theorems require
that the objective function of a learning algorithm must depend only
on $\set{\inner{A\vphi_i}{A\vphi_j}}_{i,j = 1}^n$ or equivalently
$\set{\inner{\vphi_i}{M\vphi_j}}_{i,j = 1}^n$. This condition is,
actually, not a strict condition since learners in literatures have
their objective functions in this form
\shortcite{Chen:CVPR05,Goldberger:NIPS05,Globerson:NIPS06,Weinberger:NIPS06,Yang:AAAI06,Sugiyama:ICML06,Yan:PAMI07,Wei:ICML07,Torresani:NIPS07}.
\\\\
2. Note that the two theorems stated in this section do not require
$\{\basis_i\}$ to be an orthonormal set. However, there is an
advantage of the KPCA trick which restricts $\tilde{\psi_i} =
\psi_i$ as in Eq.~\eqref{eq_newcoord};
this will be discussed in Sect.~\ref{sect_versus}.\\\\
3. A running time of each learner strongly depends on the
dimensionality of the input data. As recommended by
\citeA{Weinberger:NIPS06}, it can be helpful to first apply a
dimensionality reduction algorithm such as PCA before performing a
learning process: the learning process can be tremendously speed up
by retaining only, says, the 200 largest-variance principal
components of the input data. In the KPCA trick framework
illustrated in Figure~\ref{fig_KPCA}, dimensionality reduction can
be performed without any extra work as KPCA is already applied at
the first place.\\\\
4. The stronger version of Theorem~\ref{thm_repr2} can be achieved
by inserting a regularizer into the objective function of a
(kernelized) Mahalanobis distance learner as stated in
Theorem~\ref{thm_repr3}. For compact notations, we use the fact that
$A$ is representable by $\set{\tau_i}$ as shown in
Lemma~\ref{thm:riesz}.
\begin{thm} (Strong Representer Theorem) \label{thm_repr3}
  Define $\set{\basis_i}_{i=1}^\jdim$ and $f$ be as in Theorem~\ref{thm_repr}.
  For monotonically increasing functions $g_i$, let
  \begin{align*}
        h(\tau_1, ..., \tau_d, \phi_1, ..., \phi_n) \ = \
        f(\inner{\tau_1}{\vphi_1},\ldots,\inner{\tau_i}{\vphi_j},\ldots,\inner{\tau_n}{\vphi_d})
        + \sum_{i=1}^d g_i(\norm{\tau_i}).
  \end{align*}
  Any optimal set of linear functionals
  \begin{align*}
    & \argmin_{\{\tau_i\}} \ h(\tau_1, ..., \tau_d, \phi_1, ..., \phi_n)\\
    & \textrm{s.t.} \ \forall i \ \tau_i:\jcal{X}\to \Real \textrm{ is a bounded linear
    functional } \
  \end{align*}
  must admit the representation of
%  \begin{align*}
    $\ \ \tau_i = \sum_{j=1}^n u_{ij} \basis_j \ \ \ (i = 1, \ldots,
    d).$
%  \end{align*}
\end{thm}
The proof of this result is very similar to that of
\shortcite[Theorem 4.2]{Scholkopf:BOOK01} so that we omit its
details here. In fact, to prove the validation of KDNE, we need this
strong representer theorem (see the case of KPCA in
\shortciteA[pp.92]{Scholkopf:BOOK01}).

To apply Theorem~\ref{thm_repr3} to our framework, we can simply
view $A = (\tau_1, ..., \tau_d)^T$. If each $g_i$ is the square
function, then regularizer becomes $\sum_{i=1}^d \norm{\tau_i}^2 =
\norm{A}_{HS}$ where $\norm{\cdot}_{HS}$ is the Hilbert-Schmidt (HS)
norm of an operator. If each $\tau_i$ is finite-dimensional, the HS
norm is reduced to the Frobenius norm $\norm{\cdot}_F$. Here, we
allow the HS norm of a bounded linear operator to take a value of
$\infty$. For the kernel trick (by substituting $\basis_i =
\phi_i$), the result above states that any optimal $\{\tau_i\}$ must
be represented by $\{\Phi \mathbf{u}_i\}$. Therefore, using the same
notation as Subsection~\ref{sect_framework}, we have
\[ \sum_{i=1}^d g_i(\norm{\tau_i}) =
   \sum_{i=1}^d \norm{\tau_i}^2 =
   \sum_{i=1}^d \textbf{u}_i^T \Phi^T \Phi \textbf{u}_i =
   \sum_{i=1}^d \textbf{u}_i^T K \textbf{u}_i =
%   \mbox{trace}\big(K (\sum_{i=1}^d \textbf{u}_i\textbf{u}_i^T)\big) =
   \mbox{trace}\big(U K U^T).
\]
This regularizer is first appeared in the work of
\shortciteA{Globerson:NIPS06}. Similarly, for the KPCA trick (by
substituting $\basis_i = \psi_i$), any optimal $\{\tau_i\}$ must be
represented by $\{\Psi \mathbf{u}_i\}$ and, using the fact that
$\Psi^T \Psi = I$, we have $\sum_{i=1}^d \norm{\tau_i}^2 =
\mbox{trace}\big(U U^T)  = \norm{U}^2_F$.

By adding the regularizer, $\mbox{trace}(UKU^T)$ or $\norm{U}^2_F$,
into existing objective functions, we have a new class of learners,
namely, \emph{regularized Mahalanobis distance learners} such as
regularized KNCA (RKNCA), regularized KLMNN (RKLMNN) and regularized
KDNE (RKDNE). Our framework can be further extended into a problem
in semi-supervised settings by adding more complicated functions of
$g_i(\cdot)$ such as \emph{manifold regularizers}, see e.g.
\shortciteA{Me:Arxiv08b}. We plan to investigate effects of using
various types of regularizers in the near future.
%\section{Proof of the representer theorems} \label{sect_proof}

\section{Selection of a Kernel Function}
\label{sect_selectk} The problem of selecting an efficient kernel
function is central to all kernel machines. All previous works on
Mahalanobis distance learners use exhaustive methods such as cross
validation to select a kernel function. In this section, we
investigate a possibility to automatically construct a kernel which
is appropriate for a Mahalanobis distance learner. In the first part
of this section, we consider a popular method called \emph{kernel
alignment} \shortcite{Lanckriet:JMLR04,Zhu:NIPS05} which is able to
learn, from a training set, a kernel in the form of $k(\cdot,\cdot)
= \sum_i \alpha_i k_i(\cdot,\cdot)$ where $k_1(\cdot,\cdot), ...,
k_m(\cdot,\cdot)$ are pre-chosen base kernels. In the second part of
this section, we investigate a simple method which constructs an
unweighted combination of base kernels, $\sum_i k_i(\cdot,\cdot)$
(henceforth refered to as an \emph{unweighted kernel}). A
theoretical result is provided to support this simple approach.
Kernel constructions based on our two approaches require much
shorter running time when comparing to the standard cross validation
approach.

\subsection{Kernel Alignment} \label{sect_align}
Here, our kernel alignment formulation belongs to the class of
quadratic programs (QPs) which can be solved more efficiently than
the formulations proposed by \citeA{Lanckriet:JMLR04} and
\citeA{Zhu:NIPS05} which belong to the class of semidefinite
programs (SDPs) and quadratically constrained quadratic programs
(QCQPs), respectively  \cite{Boyd:BOOK04}.

To use kernel alignment in classification problems, the following
assumption is central: for each couple of examples $\textbf{x}_i,
\textbf{x}_j$, the ideal kernel $k(\textbf{x}_i, \textbf{x}_j)$ is
$Y_{ij}$ \shortcite{Guermeur:KMCB04} where
\begin{equation*}
Y_{ij} =
\begin{cases}
  +1, & \mbox{if } y_i = y_j,\\
  \frac{-1}{p-1}, & \mbox{otherwise},
\end{cases}
\end{equation*}
and $p$ is the number of classes in the training data. Denoting $Y$
as the matrix having elements of $Y_{ij}$, we then define the
\emph{alignment} between the kernel matrix $K$ and the ideal kernel
matrix $Y$ as follows:
\begin{equation} \label{eq_align0}
\mbox{align}(K,Y) = \frac{\langle K,Y\rangle_F}{||K||_F ||Y||_F},
\end{equation}
where $\langle\cdot,\cdot \rangle_F$ denotes the Frobenius
inner-product such that $\langle K,Y\rangle_F = \mbox{trace}(K^T Y)$
and $\norm{\cdot}_F$ is the Frobenius norm induced by the Frobenius
inner-product.

Assume that we have $m$ kernel functions, $k_1(\cdot,\cdot), ...,
k_m(\cdot,\cdot)$ and $K_1, ..., K_m$ are their corresponding Gram
matrices with respect to the training data. In this paper, the
kernel function obtained from the alignment method is parameterized
in the form of $k(\cdot,\cdot) = \sum_i \alpha_i k_i(\cdot,\cdot)$
where $\alpha_i \ge 0$. Note that the obtained kernel function is
guaranteed to be positive semidefinite. In order to learn the best
coefficients $\alpha_1, ..., \alpha_m$, we solve the following
optimization problem:
\begin{equation} \label{eq_align1}
\{\alpha_1, ..., \alpha_m\} = \arg \max_{\alpha_i \ge 0} \
\mbox{align}(K, Y),
\end{equation}
where $K = \sum_i \alpha_i K_i$. Note that as $K$ and $Y$ are PSD,
$\langle K,Y\rangle_F \ge 0$. Since both the numerator and
denominator terms in the alignment equation can be arbitrary large,
we can simply fix the numerator to 1. We then reformulate the
problem as follows:
\begin{align*}
 \underset{\alpha_i \ge 0, \langle K,Y\rangle_F = 1}{\operatorname{arg\,max}}\, \mbox{align}(K,Y)
 =& \underset{\alpha_i \ge 0, \langle K,Y\rangle_F = 1}{\operatorname{arg\,min}}\, ||K||_F ||Y||_F\\
 =&  \underset{{\alpha_i \ge 0, \langle K,Y\rangle_F = 1}}{\operatorname{arg\,min}}\, ||K||^2_F\\
 =& \underset{\alpha_i \ge 0, \sum_i \alpha_i \langle K_i,Y\rangle_F = 1}{\operatorname{arg\,min}} \sum_{i,j} \alpha_i \alpha_j \langle K_i,K_j\rangle_F.
\end{align*}
Defining a PSD matrix $S$ whose elements $S_{ij} = \langle
K_i,K_j\rangle_F$, a vector $\textbf{b} = (\langle K_1,Y\rangle_F,
..., \langle K_m,Y\rangle_F)^T$ and a vector ${\boldsymbol{\alpha}}
= (\alpha_1, ..., \alpha_m)^T$, we then reformulate Eq.~
(\ref{eq_align1}) as follows:
\begin{equation} \label{eq_align2}
{\boldsymbol{\alpha}} = \underset{\alpha_i \ge 0, \
\boldsymbol{\alpha}^T \textbf{b} = 1}{\argmin}
{\boldsymbol{\alpha}}^T S {\boldsymbol{\alpha}}.
\end{equation}
This optimization problem is a QP and can be efficiently solved
\shortcite{Boyd:BOOK04}; hence, we are able to learn the best kernel
function $k(\cdot,\cdot) = \sum_i \alpha_i k_i(\cdot,\cdot)$
efficiently.

Since the magnitudes of the optimal $\alpha_i$ are varied due to
$\norm{K_i}_F$, it is convenient to use $k'_i(\cdot, \cdot) =
k_i(\cdot, \cdot)/||K_i||_F$ and hence $K'_i = K_i/||K_i||_F$ in the
derivation of Eq.~(\ref{eq_align2}). We define $S'$ and
$\textbf{b}'$ similar to $S$ and $\textbf{b}$ except that they are
based on $K'_i$ instead of $K_i$. Let
\begin{equation} \label{eq_align3}
{\boldsymbol{\gamma}} = \underset{\gamma_i \ge 0, \
\boldsymbol{\gamma}^T \textbf{b}' = 1}{\argmin}
{\boldsymbol{\gamma}}^T S' {\boldsymbol{\gamma}}.
\end{equation}
It is easy to see that the final kernel function $k(\cdot,\cdot) =
\sum_i \gamma_i k'_i(\cdot,\cdot)$ achieved from
Eq.~(\ref{eq_align3}) is not changed from the kernel achieved from
Eq.~(\ref{eq_align2}).

Note that we can further modify Eq.~\eqref{eq_align2} to enforce
sparseness of $\boldsymbol{\alpha}$ and improve a speed of an
algorithm by minimizing an upper bound of $||K||_F$ instead of
minimizing the exact quantity so that the optimization formula
belongs to the class of linear programs (LPs) instead of QPs.
\begin{align} \label{eq_align4}
 \underset{\alpha_i \ge 0, \langle K,Y\rangle_F = 1}{\operatorname{min}}\, ||K||_F
 \le  \underset{{\alpha_i \ge 0, \langle K,Y\rangle_F = 1}}{\operatorname{min}}\, ||\mbox{vec}(K)||_1
\end{align}
where vec($\cdot$) denotes a standard ``vec'' operator converting a
matrix to a vector \cite{Minka:note97}. By using a standard trick
for an absolute-valued objective function \cite{Boyd:BOOK04},
Eq.~\eqref{eq_align4} can be solved by linear programming. Note that
the above optimization algorithm of minimizing the upper bound of a
desired objective function is similar to the popular support vector
machines where the hinge loss is minimized instead of the 0/1 loss.

\subsection{Unweighted Kernels} \label{sect_unweighted}
In this subsection, we show that a very simple kernel $k'(\cdot,
\cdot) = \sum_i k_i(\cdot, \cdot)$ is theoretically efficient, no
less than a kernel obtained from the alignment method. Denote
$\phi_i^k$ as a mapped vector of an original example $\textbf{x}_i$
by a map associated with a kernel $k(\cdot,\cdot)$. The main idea of
the contents presented in this section is the following simple but
useful result.

\begin{prp} \label{thm_unweighted}
  Let $\{\alpha_i\}$ be a set of positive
  coefficients, $\alpha_i > 0$ for each $i$, and let $k_1(\cdot,\cdot), ...,
  k_m(\cdot,\cdot)$ be base PSD kernels and $k(\cdot, \cdot) = \sum_i \alpha_i k_i(\cdot, \cdot)$
  and $k'(\cdot, \cdot) = \sum_i k_i(\cdot, \cdot)$. Then, there
  exists an invertible linear map $B$ such that $B: \phi_i^{k'} \rightarrow \phi_i^k$ for each $i$.
\end{prp}
\begin{proof}
Without loss of generality, we will concern here only the case of $m
= 2$; the cases such that $m > 2$ can be proven by induction. Let
$\jcal{H}_i \oplus \jcal{H}_j$ be a direct sum of $\jcal{H}_i$ and
$\jcal{H}_j$ where its inner product is defined by
$\inner{\cdot}{\cdot}_{\jcal{H}_i}+\inner{\cdot}{\cdot}_{\jcal{H}_j}$
and let $\{\phi_i^{(j)}\} \subset \jcal{H}_j$ denote a mapped
training set associated with the $j^{th}$ base kernel. Then we can
view $\phi_i^k = (\sqrt{\alpha_1}\phi_i^{(1)},
\sqrt{\alpha_2}\phi_i^{(2)}) \in \jcal{H}_i \oplus \jcal{H}_j$ since
\begin{align*}
  \inner{\phi_i^k}{\phi_j^k} = k(\textbf{x}_i,\textbf{x}_j)
  &= \alpha_1 k_1(\textbf{x}_i,\textbf{x}_j) + \alpha_2 k_2(\textbf{x}_i,\textbf{x}_j)\\
  &= \inner{\sqrt{\alpha_1}\phi^{(1)}_i}{\sqrt{\alpha_1}\phi_j^{(1)}} \ + \inner{\sqrt{\alpha_2}\phi_i^{(2)}}{\sqrt{\alpha_2}\phi_j^{(2)}}\\
  &= \inner{\left(\sqrt{\alpha_1}\phi^{(1)}_i, \sqrt{\alpha_2}\phi^{(2)}_i \right)}{\left(\sqrt{\alpha_1}\phi^{(1)}_j, \sqrt{\alpha_2}\phi^{(2)}_j\right)
  }.
\end{align*}
Similarly, we can also view $\phi_i^{k'} = (\phi_i^{(1)},
\phi_i^{(2)}) \in \jcal{H}_i \oplus \jcal{H}_j$. Let $I_j$ be the
identity map in ${\mathcal{H}}_j$. Then,
\[B =
    \begin{bmatrix}
     \sqrt{\alpha_1}I_1 & 0 \\
     0 & \sqrt{\alpha_2}I_2
    \end{bmatrix}.
\]
Since $\infty > \alpha_1, \alpha_2 > 0$ and $B$ is bounded (the
operator norm of $B$ is max$(\sqrt{\alpha_1},\sqrt{\alpha_2})$), $B$
is invertible.
\end{proof}

Now suppose we apply the kernel $k(\cdot, \cdot) = \sum_i \alpha_i
k_i(\cdot, \cdot)$ obtained from the kernel alignment method to a
Mahalanobis distance learner and an optimal transformation $A^*$ is
returned. Let $f(\cdot)$ be an objective function which depends only
on an inner product $\langle A\phi_i , A\phi_j\rangle$ (as assumed
in Theorems~\ref{thm_repr} and ~\ref{thm_repr2}). Since, from
Proposition~\ref{thm_unweighted}, $\inner{A^*\phi^k_i}{A^*\phi^k_j}
= \inner{A^*B\phi^{k'}_i}{A^*B\phi^{k'}_j}$, we have
\[
f^* \equiv
f\left(\left\{\inner{A^*\phi^k_i}{A^*\phi^k_j}\right\}\right) =
f\left(\left\{\inner{A^*B\phi^{k'}_i}{A^*B\phi^{k'}_j}\right\}\right).
\]
Thus, by applying a training set $\{\phi^{k'}_i\}$ to a learner who
tries to minimize $f(\cdot)$, a learner will return a linear map
with the objective value less than or equal to $f^*$ (because the
learner can at least return $A^*B$). Notice that because $B$ is
invertible, the value $f^*$ is in fact optimal. Consequently, the
following claim can be stated: ``there is no need to apply the
methods which learn $\{\alpha_i\}$, e.g. the kernel alignment
method, \emph{at least in theory}, because learning with a simple
kernel $k'(\cdot, \cdot)$ also results in a linear map having the
same optimal objective value''. However, \emph{in practice}, there
can be some differences between using the two kernels
$k(\cdot,\cdot)$ and $k'(\cdot,\cdot)$ due to the following reasons.

$\bullet$ \textbf{Existence of a local solution}. As some
optimization problems are not convex, there is no guarantee that a
solver is able to discover a global solution within a reasonable
time. Usually, a learner discovers only a local solution, and hence
two learners based on $k(\cdot,\cdot)$ and $k'(\cdot,\cdot)$ will
not give the same solution. KNCA belongs to this case.

$\bullet$ \textbf{Non-existence of the unique global solution}. In
some optimization problems, there can be many different linear maps
having the same optimal values $f^*$, and hence there is no
guarantee that two learners based on $k(\cdot,\cdot)$ and
$k'(\cdot,\cdot)$ will give the same solution. KLMNN is an example
of this case.

$\bullet$ \textbf{Size constraints}. Because of a size constraint
such as $A A^T = I$ used in KDNE, our arguments used in the previous
subsection cannot be applied, i.e., given that $A^{*T} A^* = I$,
there is no guaranteed that $(A^*B) (A^*B)^T = I$. Hence, $A^*B$ may
not be an optimal solution of a learner based on $k'(\cdot,\cdot)$.

$\bullet$ \textbf{Preprocessing of target neighbors}. The behavior
of some learners depends on their preprocesses. For example, before
learning takes place, the KLMNN and KDNE algorithms have to specify
the target neighbors of each point (by specifying a value of
$w_{ij}$). In a case of using the KPCA trick, this specification is
based on the Euclidean distance with respect to a selected kernel
(see Subsection~\ref{sect_klmnn} and
Proposition~\ref{thm_parseval}). In this case, the Euclidean
distance with respect to an aligned kernel $k(\cdot,\cdot)$ (which
already used some information of a training set) is more appropriate
than the Euclidean distance with respect to an unweighted kernel
$k'(\cdot,\cdot)$.

$\bullet$ \textbf{Zero coefficients}.  In the above proposition we
assume $\alpha_i > 0$ for all $i$. Often, the alignment algorithm
return $\alpha_i = 0$ for some $i$. Define $A^*$ and $f^*$ as above.
Following the same line of the proof of
Proposition~\ref{thm_unweighted}, in the cases that the alignment
method gives $\alpha_i = 0$ for some $i$, it can be easily shown
that a learner with a kernel $k'(\cdot,\cdot)$ will return a linear
map with its objective value better than or equal to $f^*$.
%Nevertheless, note that, in machine learning, a better value of an
%inappropriate objective functiona can lead to
%overfitting.\\

Since constructing $k'(\cdot,\cdot)$ is extremely easy,
$k'(\cdot,\cdot)$ is a very attractive choice to be used in
kernelized algorithms.

\section{Demonstrations}
In this section, the advantages of the KPCA trick over the kernel
trick are demonstrated. After that, we conduct extensive experiments
to illustrate the performance of kernelized algorithms, especially
for those applying the kernel construction methods described in the
previous section.

\subsection{KPCA Trick versus Kernel Trick} \label{sect_versus}
To understand the advantages of the KPCA trick over the kernel
trick, it is best to derive a kernel trick formula for each
algorithm and see what have to be done in order to implement a
kernelized algorithm applying the kernel trick. In this section, we
define $\{\phi_i\}$ and $\Phi$ as in Section~\ref{sect_framework}.

\subsubsection{KNCA}
As noted in Sect.~\ref{sect_nca}, in order to minimize the objective
of NCA and KNCA, we need to derive gradient formulas, and the
formula of $\partial f^{KNCA}/\partial A$ is
\cite{Goldberger:NIPS05}:
\begin{equation} \label{eq_gradknca}
-2A \sum_i \left(p_i \sum_k p_{ik}\phi_{ik}\phi_{ik}^T - \sum_{j \in
c_i} p_{ij}\phi_{ij}\phi_{ij}^T \right)
\end{equation}
where for brevity we denote $\phi_{ij} = \phi_i - \phi_j$.
Nevertheless, since $\phi_i$ may lie in an infinite dimensional
space, the above formula cannot be always implemented in practice.
In order to implement the kernel-trick version of KNCA, users need
to prove the following proposition which is not stated in the
original work of \citeA{Goldberger:NIPS05}:
\begin{prp} \label{thm_knca}
$\partial f^{KNCA}/\partial A$ can be formulated as $V \Phi^T$ where
$V$ depends on $\{\phi_i\}$ only in the form of
$\inner{\phi_i}{\phi_j} = k(\textbf{x}_i,\textbf{x}_j)$, and thus we
can compute all elements of $V$.
\end{prp}
\begin{proof}
Define a matrix $B^\phi_i = (0, 0, ..., \phi,..., 0, 0)$ as a matrix
with its $i^{th}$ column is $\phi$ and zero vectors otherwise.
Denote $\textbf{k}_{ij} = \textbf{k}_{i} - \textbf{k}_{j}$.
Substitute $A = U\Phi^T$ to Eq.~\eqref{eq_gradknca} we have
\begin{align*}
\frac{\partial f^{KNCA}}{\partial A} &= -2U \sum_i \Big(p_i \sum_k
                                       p_{ik}\textbf{k}_{ik}\phi_{ik}^T - \sum_{j \in c_i}
                                       p_{ij}\textbf{k}_{ij}\phi_{ij}^T \Big)\\
         &= -2U \sum_i \Big(p_i \sum_k p_{ik} (B^{\textbf{k}_{ik}}_i - B^{\textbf{k}_{ik}}_k) \ \ -\\
         & \ \quad \quad \quad \sum_{j \in c_i} p_{ij}(B^{\textbf{k}_{ij}}_i - B^{\textbf{k}_{ij}}_j) \Big) \Phi^T\\
         &= \ \ V \Phi^T,
\end{align*}
which completes the proof.
\end{proof}
Therefore, at the $i^{th}$ iteration of an optimization step of a
gradient optimizer, we needs to update the current best linear map
as follows: \noindent
\begin{align} \label{eq_knca}
 A^{(i)} = A^{(i-1)} + \epsilon\frac{\partial f^{KNCA}}{\partial A}
 &= (U^{(i-1)} + \epsilon V^{(i-1)})\Phi^T \nonumber\\
 &= U^{(i)} \Phi^T,
\end{align}
where $\epsilon$ is a step size. The kernel-trick formulas of KNCA
are thus finally achieved. However, we emphasize that the process of
proving Proposition~\ref{thm_knca} and Eq.~\eqref{eq_knca} is not
trivial and may be tedious and difficult for non-experts as well as
practitioners who focus their tasks on applications rather than
theories. Moreover, since the formula of $\partial f^{KNCA}/\partial
A$ is significantly different from $\partial f^{NCA}/\partial A$,
users are required to re-implement KNCA (even they already possess
an NCA implementation) which is again not at all convenient. In
contrast, we note that all these difficulties are disappeared if the
KPCA trick algorithm consisting of three simple steps shown in
Fig.~\ref{fig_KPCA} is applied instead of the kernel trick.

There is another advantage of using the KPCA trick on
KNCA\footnote{We slightly modify the code of Charless Fowlkes:
\textsf{http://www.cs.berkeley.edu/$\sim$fowlkes/software/nca/}}. By
the nature of a gradient optimizer, it takes a large amount of time
for NCA and KNCA to converge to a local solution, and thus a method
of speeding up the algorithms is needed. As recommended by
\citeA{Weinberger:NIPS06}, it can be helpful to first apply PCA
before performing a learning process: the learning process can be
tremendously speed up by retaining only, says, the 100
largest-variance principal components of the input data. In the KPCA
trick framework, no extra work is required for this speed-up task as
KPCA is already applied at the first place.

\subsubsection{KLMNN} \label{sect_klmnn}
Similar to KNCA, the online-available code of LMNN\footnote{
\textsf{http://www.weinbergerweb.net/Downloads/LMNN.html}} employs a
gradient based optimization, and thus new gradient formulas in the
feature space has to be derived and new implementation has to be
done in order to apply the kernel trick. On the other hand, by
applying the KPCA trick, the original LMNN code can be immediately
used.

There is another advantage of the KPCA trick on LMNN: LMNN requires
a specification of $w_{ij}$ which is usually based on the quantity
$\norm{\textbf{x}_i - \textbf{x}_j}$. Thus, it makes sense that
$w_{ij}$ should be based on $\norm{\phi_i - \phi_j} =
\sqrt{k(\textbf{x}_i,\textbf{x}_i) + k(\textbf{x}_j,\textbf{x}_j) -
2k(\textbf{x}_i,\textbf{x}_j)}$ with respect to the feature space of
KLMNN, and hence, with the kernel trick, users have to modify the
original code in order to appropriately specify $w_{ij}$. In
contrast, by applying the KPCA trick which restricts $\{\psi_i\}$ to
be an orthonormal set as in Eq.~\eqref{eq_newcoord}, we have the
following proposition:
\begin{prp} \label{thm_parseval} Let $\set{\psi_i}_{i=1}^\jdim$ be an orthonormal set
  such that $span(\set{\psi_i}_{i=1}^\jdim) = span(\set{\phi_i}_{i=1}^\jdim)$
  and
  $\tphi_i=\left(\inner{\vphi_i}{\psi_1},\ldots,\inner{\vphi_i}{\psi_\jdim}\right)^T
  \in \Real^\jdim$, then
  $\norm{\tphi_i - \tphi_j}^2 = \norm{\phi_i~-~\phi_j}^2$ for each
  $1 \le i,j \le n$.
\end{prp}
\begin{proof} Since we work on a separable Hilbert space $\jcal{X}$, we can
extend the orthonormal set $\{\psi_i\}_{i=1}^\jdim$ to
$\{\psi_i\}_{i=1}^\infty$ such that
$\overline{span(\{\psi_i\}_{i=1}^\infty)}$ is $\jcal{X}$ and
$\inner{\phi_i}{\psi_j} = 0$ for each $i = 1, ..., n$ and $j >
\jdim$. Then, by an application of the Parseval identity
\cite{Wicharn:BOOK06},
\begin{align*}
  \norm{\phi_i - \phi_j}^2
  = \sum_{k=1}^\infty \langle \phi_i-\phi_j ,\psi_k
    \rangle^2
  &= \sum_{k=1}^\jdim \langle \phi_i-\phi_j ,\psi_k
    \rangle^2\\
  &= \norm{\tphi_i - \tphi_j}^2.
\end{align*}
The last equality comes from Eq.\eqref{eq_newcoord}.
\end{proof}
Therefore, with the KPCA trick, the target neighbors $w_{ij}$ of
each point is computed based on $\norm{\tphi_i - \tphi_j} =
\norm{\phi_i - \phi_j}$ without any modification of the original
code.
\subsubsection{KDNE}
By applying $A = U \Phi^T$ and defining the gram matrix $K = \Phi^T
\Phi$, we have the following proposition.
\begin{prp}
The kernel-trick formula of KDNE is the following minimization
problem:
\begin{equation} \label{eq_kdne}
  U^* = \underset{U K U^T = I}{\operatorname{\argmin}}\, \mbox{trace}(UK(D-W)KU^T).
\end{equation}
\end{prp}
\noindent Note that this kernel-trick formula of KDNE involves a
\emph{generalized eigenvalue problem} instead of a plain eigenvalue
problem involved in DNE. As a consequence, we face a
\emph{singularity} problem, i.e. if $K$ is not full-rank, the
constraint $U K U^T = I$ cannot be satisfied. Using elementary
linear algebra, it can be shown that $K$ is not full-rank if and
only if $\{\phi_i\}$ is not linearly independent, and this condition
is not highly improbable. \citeA{Sugiyama:ICML06}, \citeA{Yu:PR01},
and \citeA{Yang:PR03} suggest methods to cope with the singularity
problem in the context of Fisher discriminant analysis which may be
applicable to KDNE. \citeA{Sugiyama:ICML06} recommends to use the
constraint $U (K +\epsilon I) U^T = I$ instead of the original
constraint; however, an appropriate value of $\epsilon$ has to be
tuned by cross validation which is time-consuming. Alternatively,
\citeA{Yu:PR01} and \citeA{Yang:PR03} propose more complicated
methods of directly minimizing an objective function in the null
space of the constraint matrix so that the singularity problem is
explicitly avoided.

We note that a KPCA-trick implementation of KDNE does not have this
singularity problem as only a plain eigenvalue problem has to be
solved. Moreover, as in KLMNN, applying the KPCA trick instead of
the kernel trick to KDNE avoid the tedious task of modifying the
original code to appropriately specify $w_{ij}$ in the feature
space.

\subsection{Numerical Experiments} \label{sect_exp} 
\begin{figure*}[t]  \begin{center} \vskip -0.6in \hbox{\hskip 1.20in
\setlength{\epsfxsize}{1.75in} \epsfbox{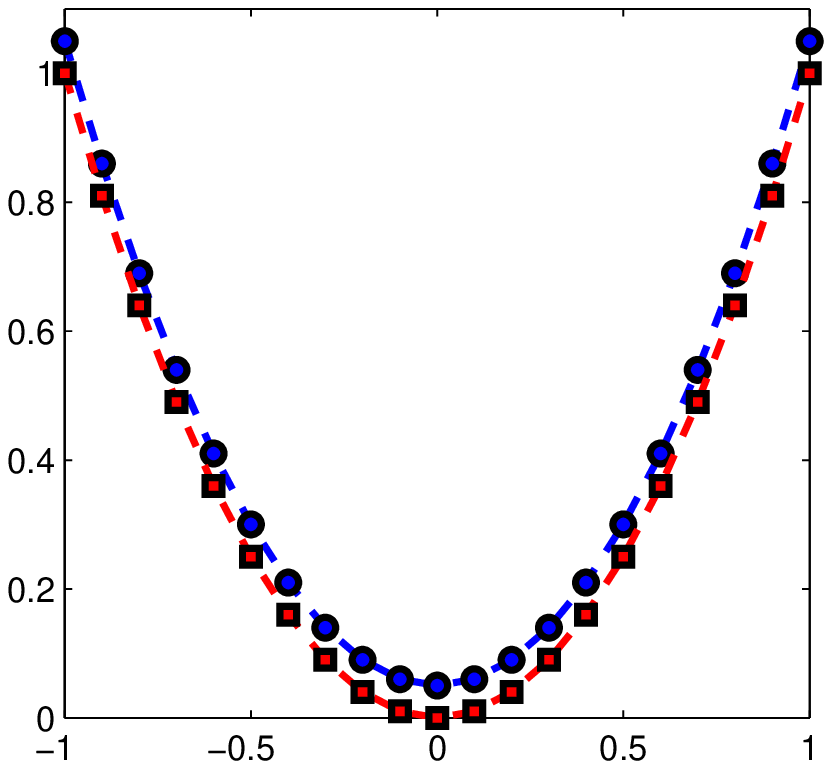}
\setlength{\epsfxsize}{1.75in} \epsfbox{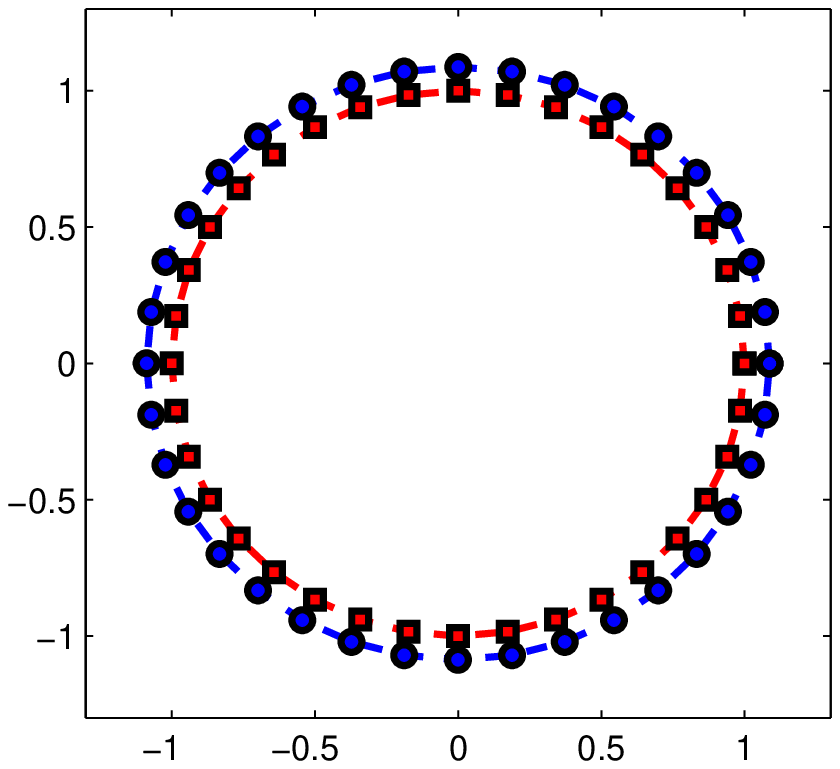} }
\end{center}
\vskip -0.6in \caption{ Two synthetic examples where NCA, LMNN and
DNE cannot learn any efficient Mahalanobis distances for kNN. Note
that in each example, data in each class lie on a simple non-linear
1-dimensional subspace (which, however, cannot be discovered by the
three learners). In contrast, the kernel versions of the three
algorithms (using the $2^{nd}$-order polynomial kernel) can learn
very efficient distances, i.e., the non-linear subspaces are
discovered by the kernelized algorithms.} \label{fig_synthetic}
\vskip -0.3in
\end{figure*}

On page 8 of the
LMNN paper \shortcite{Weinberger:NIPS06}, Weinberger et al. gave a
comment about KLMNN: \emph{`as LMNN already yields highly nonlinear
decision boundaries in the original input space, however, it is not
obvious that ``kernelizing'' the algorithm will lead to significant
further improvement'}. Here, before giving experimental results, we
explain why ``kernelizing'' the algorithm can lead to significant
improvements. The main intuition behind the kernelization of
``Mahalanobis distance learners for the kNN classification
algorithm'' lies in the fact that non-linear boundaries produced by
kNN (with or without Mahalanobis distance) is usually helpful for
problems with multi-modalities; however, the non-linear boundaries
of kNN is sometimes not helpful when data of the same class stay on
a low-dimensional non-linear manifold as shown in
Figure~\ref{fig_synthetic}.

%In this section, we divide our experiments into two parts.
%In Subsection~\ref{sect_accur1},
In this section, we conduct experiments on NCA, LMNN, DNE and their
kernel versions on nine real-world datasets to show that (1) it is
really the case that the kernelized algorithms usually outperform
their original versions on real-world datasets, and (2) the
performance of linearly combined kernels achieved by the two methods
presented in Section~\ref{sect_selectk} are comparable to kernels
which are exhaustively selected, but the kernel alignment method
requires much shorter running time.
%In
%Subsection~\ref{sect_bivar}, we carefully investigate the
%performance of linearly combined kernels by using the
%\emph{bias-variance} analysis \cite{James:mlj03,Domingos:icml00}.

%\subsection{Generalization Performance}
%\label{sect_accur1}
%[add figures?]; for multimodality.

\begin{table}[t] %\vskip -0.30in
\caption{The average accuracy with standard deviation of NCA and
their kernel versions. On the bottom row, the win/draw/lose
statistics of each kernelized algorithm comparing to its original
version is drawn.} \label{table_nca}
\begin{center}
\vskip -0.15in
\begin{small}
\begin{sc}
\vskip -0.25in
\begin{tabular}{l||c||c||c||c}
\hline
%\abovespace\belowspace
Name & NCA & KNCA & Aligned KNCA & Unweighted KNCA\\
\hline
%\abovespace
Balance          & 0.89 $\pm$ 0.03  & \textbf{0.92 $\pm$ 0.01}& \textbf{0.92 $\pm$ 0.01} & \textbf{0.91 $\pm$ 0.03}\\
Breast Cancer    & 0.95 $\pm$ 0.01 & \textbf{0.97 $\pm$ 0.01} & \textbf{0.96 $\pm$ 0.01} & \textbf{0.96 $\pm$ 0.02}\\
Glass            & 0.61 $\pm$ 0.05 & \textbf{0.69 $\pm$ 0.02} & \textbf{0.69 $\pm$ 0.04} & \textbf{0.68 $\pm$ 0.04}\\
Ionosphere       & 0.83 $\pm$ 0.04 & \textbf{0.94 $\pm$ 0.03} & \textbf{0.92 $\pm$ 0.02} & \textbf{0.90 $\pm$ 0.03}\\
Iris             & 0.96 $\pm$ 0.03 & 0.96 $\pm$ 0.01 & 0.95 $\pm$ 0.03 & 0.96 $\pm$ 0.02\\
Musk2            & 0.87 $\pm$ 0.02 & \textbf{0.90 $\pm$ 0.01} & \textbf{0.88 $\pm$ 0.02} & 0.87 $\pm$ 0.02\\
Pima             & 0.68 $\pm$ 0.02 & \textbf{0.71 $\pm$ 0.02} & 0.67 $\pm$ 0.03 & \textbf{0.69 $\pm$ 0.01}\\
Satellite        & 0.82 $\pm$ 0.02 & \textbf{0.84 $\pm$ 0.01} & \textbf{0.84 $\pm$ 0.01} & 0.82 $\pm$ 0.02\\
Yeast            & 0.47 $\pm$ 0.02 & \textbf{0.50 $\pm$ 0.01} & \textbf{0.49 $\pm$ 0.02} & 0.47 $\pm$ 0.02\\
\hline
%\belowspace
Win/Draw/Lose   &       -          &    8/1/0              & 7/0/2                    & 5/4/0\\
%\belowspace
\hline

\end{tabular}
\end{sc}
\end{small}
\end{center}
\vskip -0.20in
\end{table}

\begin{table}[t] \vskip -0.10in\caption{The average accuracy with standard deviation of LMNN and their
kernel versions.} \label{table_lmnn}
\begin{center}
\begin{small}
\vskip -0.20in
\begin{sc}
\vskip -0.20in
\begin{tabular}{l||c||c||c||c}
\hline
%\abovespace\belowspace
Name &  LMNN & KLMNN & Aligned KLMNN & Unweighted KLMNN\\
\hline
%\abovespace
Balance          & 0.84  $\pm$ 0.04 & \textbf{0.87 $\pm$ 0.01} & \textbf{0.88 $\pm$ 0.02} & \textbf{0.85 $\pm$ 0.01}\\
Breast Cancer    & 0.95  $\pm$ 0.01 & \textbf{0.97 $\pm$ 0.01} & \textbf{0.97 $\pm$ 0.00} & \textbf{0.97 $\pm$ 0.00}\\
Glass            & 0.63  $\pm$ 0.05 & \textbf{0.69 $\pm$ 0.04} & \textbf{0.69 $\pm$ 0.04} & \textbf{0.66 $\pm$ 0.05}\\
Ionosphere       & 0.88  $\pm$ 0.02 & \textbf{0.95 $\pm$ 0.02} & \textbf{0.94 $\pm$ 0.02} & \textbf{0.94 $\pm$ 0.02}\\
Iris             & 0.95  $\pm$ 0.02 & \textbf{0.96 $\pm$ 0.02} & 0.95 $\pm$ 0.02 & \textbf{0.97 $\pm$ 0.01}\\
Musk2            & 0.80  $\pm$ 0.03 & \textbf{0.93 $\pm$ 0.01} & \textbf{0.88 $\pm$ 0.02} & \textbf{0.86 $\pm$ 0.02}\\
Pima             & 0.68  $\pm$ 0.02 & \textbf{0.71 $\pm$ 0.02} & \textbf{0.72 $\pm$ 0.02} & 0.67 $\pm$ 0.03\\
Satellite        & 0.81  $\pm$ 0.01 & \textbf{0.85 $\pm$ 0.01} & \textbf{0.84 $\pm$ 0.01} & \textbf{0.83 $\pm$ 0.02}\\
Yeast            & 0.47  $\pm$ 0.02 & \textbf{0.48 $\pm$ 0.02} & \textbf{0.54 $\pm$ 0.02} & \textbf{0.50 $\pm$ 0.02}\\
\hline
%\belowspace
Win/Draw/Lose   &       -          &    9/0/0              & 8/1/0                    & 8/0/1\\
%\belowspace
\hline
\end{tabular}
\end{sc}
\end{small}
\end{center}
%\vskip -0.3in
\end{table}

\begin{table}[t] \vskip -0.30in
\caption{The average accuracy with standard deviation of DNE  and
their kernel versions.} \label{table_dne}
\begin{center}
\vskip -0.10in
\begin{small}
\begin{sc}
\vskip -0.20in
\begin{tabular}{l||c||c||c||c}
\hline
%\abovespace\belowspace
Name & DNE & KDNE & Aligned KDNE & Unweighted KDNE\\
\hline
%\abovespace
Balance          & 0.79 $\pm$ 0.02 & \textbf{0.90 $\pm$ 0.01} & \textbf{0.83 $\pm$ 0.02} & \textbf{0.85 $\pm$ 0.03}\\
Breast Cancer    & 0.96 $\pm$ 0.01 & \textbf{0.97 $\pm$ 0.01} & 0.96 $\pm$ 0.01 & 0.96 $\pm$ 0.02\\
Glass            & 0.65 $\pm$ 0.04 & \textbf{0.70 $\pm$ 0.03} & \textbf{0.69 $\pm$ 0.04} & 0.65 $\pm$ 0.03\\
Ionosphere       & 0.87 $\pm$ 0.02 & \textbf{0.95 $\pm$ 0.02} & \textbf{0.95 $\pm$ 0.02} & \textbf{0.93 $\pm$ 0.03}\\
Iris             & 0.95 $\pm$ 0.02 & \textbf{0.97 $\pm$ 0.02} & \textbf{0.96 $\pm$ 0.02} & \textbf{0.96 $\pm$ 0.03}\\
Musk2            & 0.89 $\pm$ 0.02 & \textbf{0.91 $\pm$ 0.01} & 0.89 $\pm$ 0.02 & 0.84 $\pm$ 0.03\\
Pima             & 0.67 $\pm$ 0.02 & \textbf{0.69 $\pm$ 0.02} & \textbf{0.70 $\pm$ 0.03} & \textbf{0.70 $\pm$ 0.02}\\
Satellite        & 0.84 $\pm$ 0.01 & \textbf{0.85 $\pm$ 0.01} & \textbf{0.85 $\pm$ 0.01} & 0.81 $\pm$ 0.02\\
Yeast            & 0.40 $\pm$ 0.05 & \textbf{0.48 $\pm$ 0.01} & \textbf{0.47 $\pm$ 0.04} & \textbf{0.52 $\pm$ 0.02}\\
%\belowspace
\hline
%\belowspace
Win/Draw/Lose   &       -          &    9/0/0              & 7/2/0                    & 5/2/2\\
\hline
\end{tabular}
\end{sc}
\end{small}
\end{center}
\vskip -0.3in
\end{table}

\begin{figure}[h]
\begin{center}
\vskip -0.7in \setlength{\epsfxsize}{3.72in}
\centerline{\epsfbox{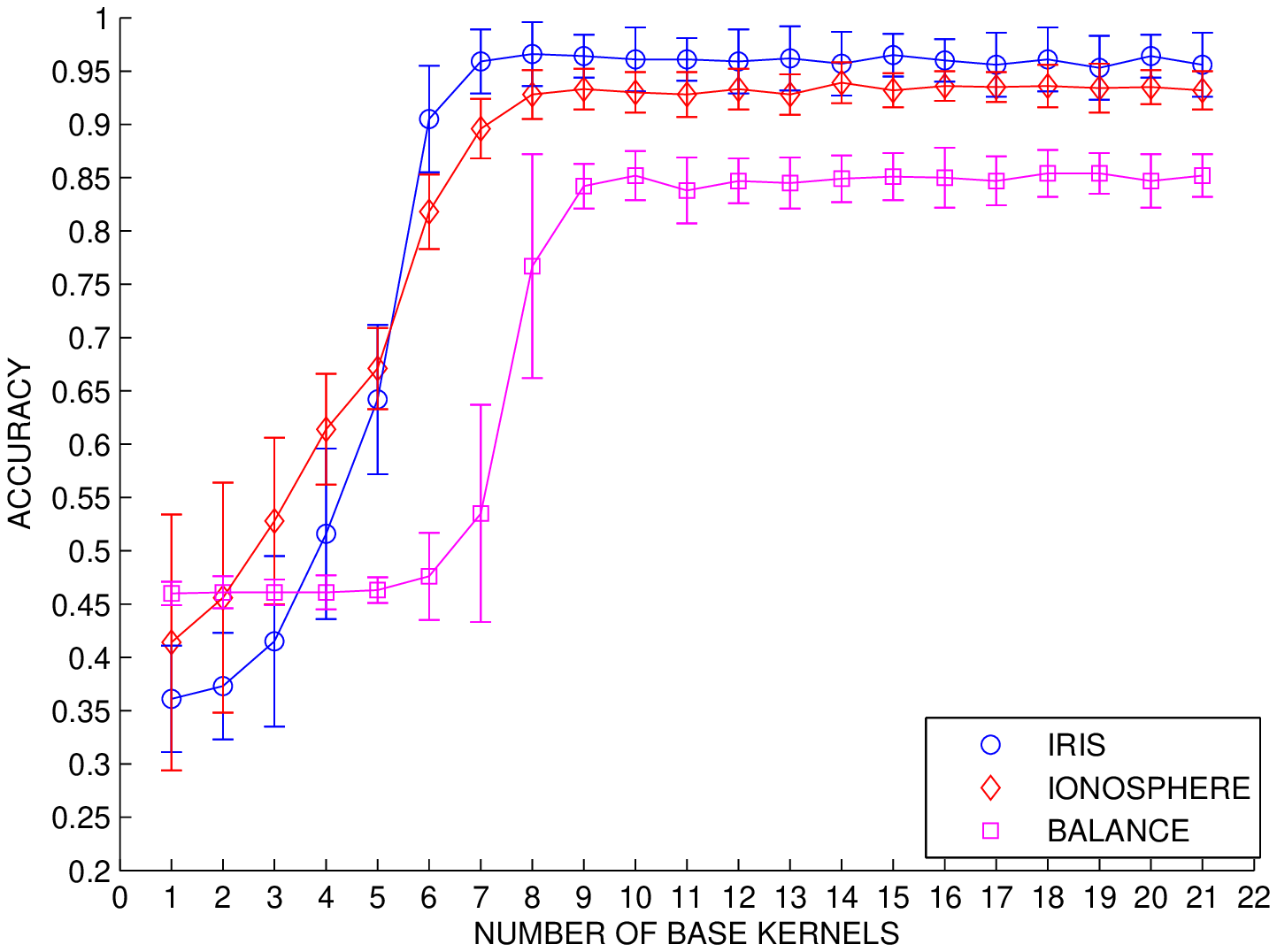}}
\end{center}
\begin{center}
\vskip -0.45in \setlength{\epsfxsize}{3.72in}
\centerline{\epsfbox{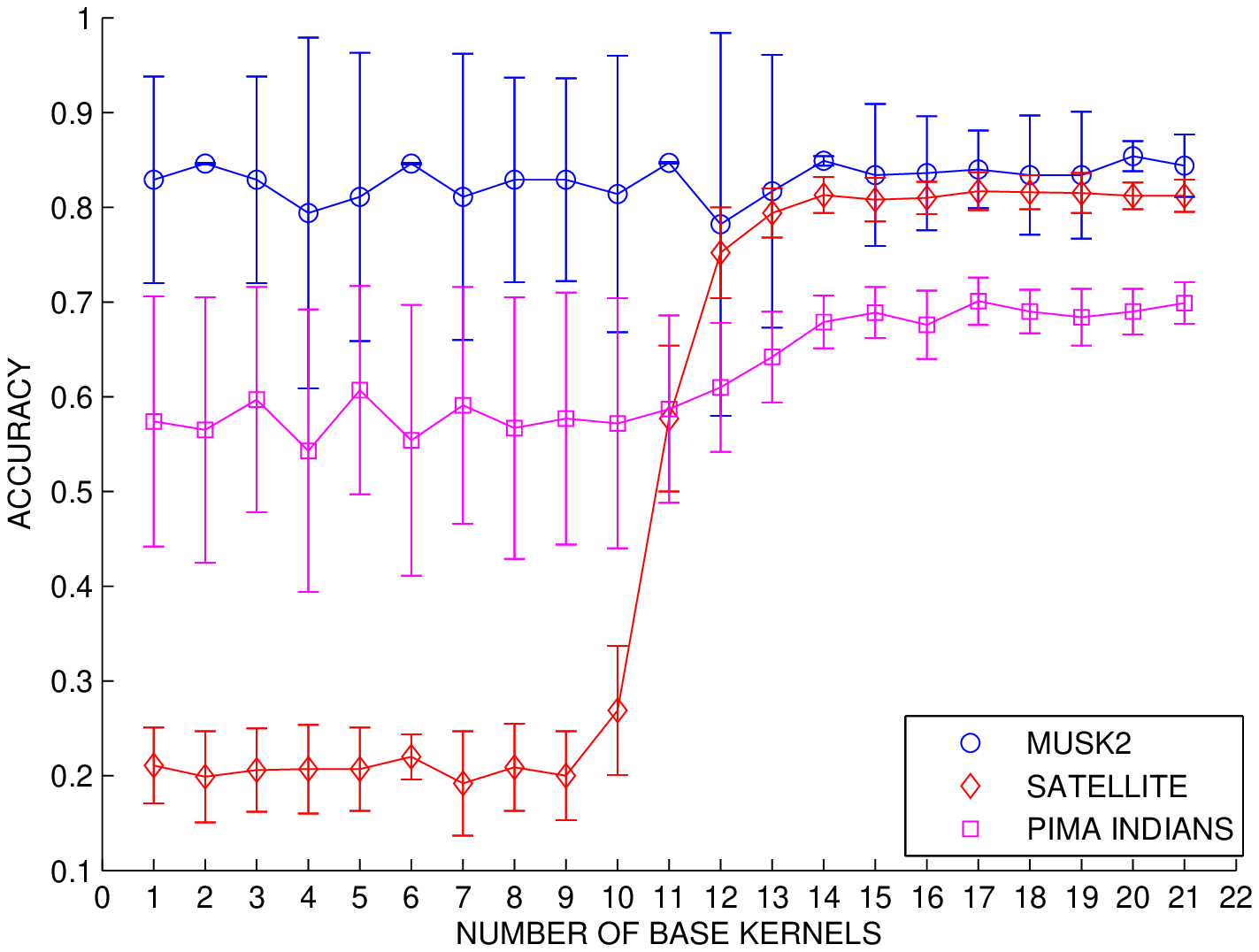}}
\end{center}
\begin{center}
\vskip -0.45in \setlength{\epsfxsize}{3.72in}
\centerline{\epsfbox{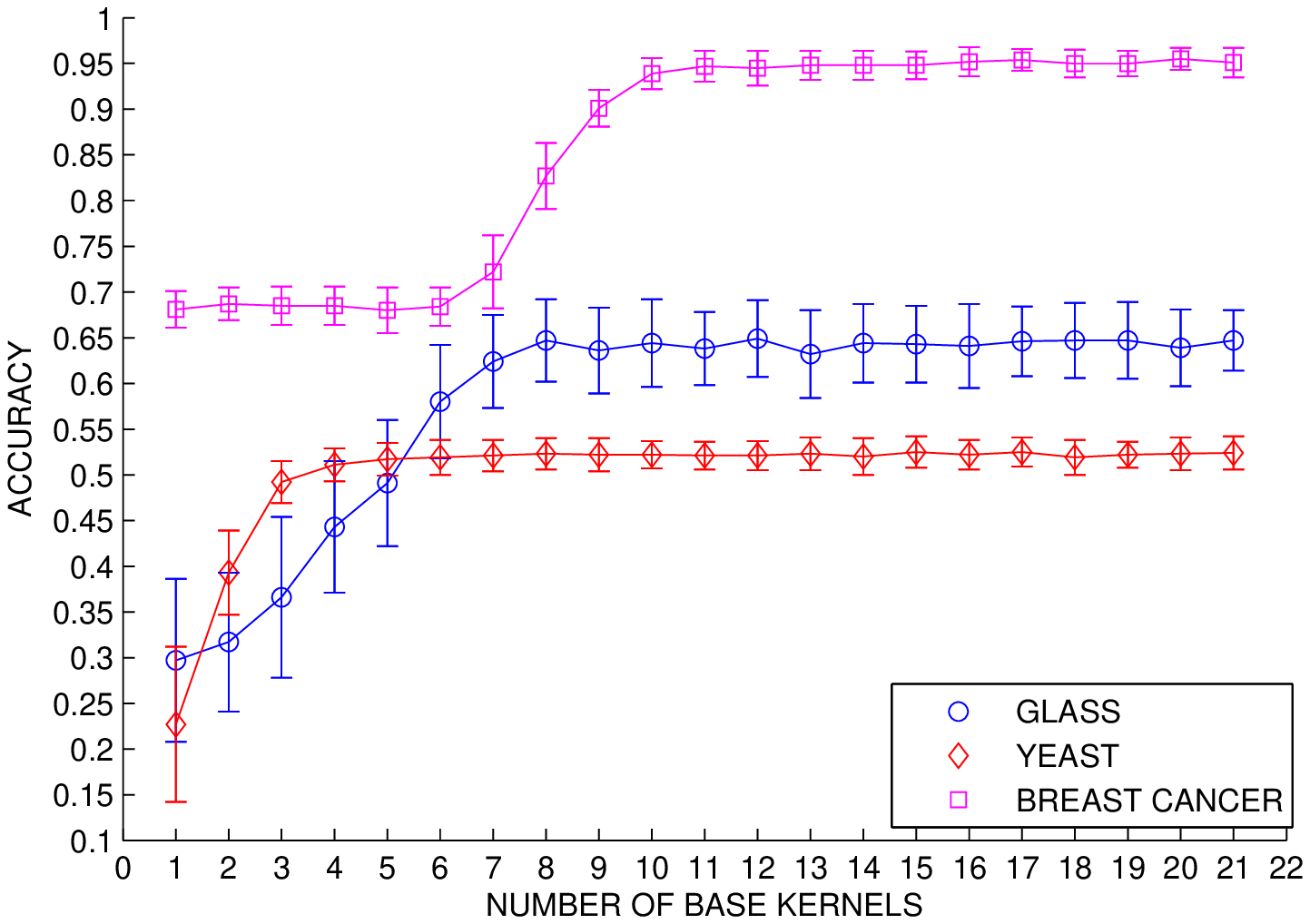}} \vskip -0.2in \caption{This
figure illustrates performance of \textsc{Unweighted KDNE} with
different number of base kernels. It can be observed from the figure
that the generalization performance of \textsc{Unweighted KDNE} will
be eventually stable as we add more and more base kernels.}
\label{fig_overfit}
\end{center} \vskip -0.2in
\end{figure}

To measure the generalization performance of each algorithm, we use
the nine real-world datasets obtained from the UCI repository
\cite{data:UCI}: \textsc{Balance}, \textsc{Breast Cancer},
\textsc{Glass}, \textsc{Ionosphere}, \textsc{Iris}, \textsc{Musk2},
\textsc{Pima}, \textsc{Satellite} and \textsc{Yeast}. Following
previous works, we randomly divide each dataset into training and
testing sets. By repeating the process 40 times, we have 40 training
and testing sets for each dataset. The generalization performance of
each algorithm is then measured by the average test accuracy over
the 40 testing sets of each dataset. The number of training data is
200 except for \textsc{Glass} and \textsc{Iris} where we use 100
examples because these two datasets contain only 214 and 150 total
examples, respectively.

Following previous works, we use the 1NN classifier in all
experiments. In order to kernelize the algorithms, three approaches
are applied to select appropriate kernels:

$\bullet$ Cross validation (\textsc{KNCA}, \textsc{KLMNN} and
\textsc{KDNE}).

$\bullet$ Kernel alignment (\textsc{Aligned KNCA}, \textsc{Aligned
KLMNN} and \textsc{Aligned KDNE}).

$\bullet$ Unweighted combination of base kernels (\textsc{Unweighted
KNCA}, \textsc{Unweighted
KLMNN} and \textsc{Unweighted KDNE}).\\\\
For all three methods, we consider scaled RBF base kernels
\shortcite[p. 216]{Scholkopf:BOOK01}, $k(x,y) =
\exp(-\frac{\norm{x-y}^2}{2 D \sigma^2})$ where $D$ is the
dimensionality of input data. Twenty one based kernels specified by
the following values of $\sigma$ are considered: 0.01, 0.025, 0.05,
0.075, 0.1, 0.25, 0.5, 0.75, 1, 2.5, 5, 7.5, 10, 25, 50, 75, 100,
250, 500, 750, 1000. all kernelized algorithms are implemented by
the KPCA trick illustrated in Figure~\ref{fig_KPCA}. As noted in
Subsection~\ref{sect_unweighted}, the main problem of using the
unweighted kernel to algorithms such as KLMNN and KDNE is that the
Euclidean distance with respect to the unweighted kernel is not
informative and thus should not be used to specify target neighbors
of each point. Therefore, in cases of KLMNN and KDNE which apply the
unweighted kernel, we employ the Euclidean distance with respect to
the input space to specify target neighbors. We slightly modify the
original codes of LMNN and DNE to fulfill this desired
specification.

The experimental results are shown in Tables~\ref{table_nca},
\ref{table_lmnn} and \ref{table_dne}. From the results, it is clear
that the kernelized algorithms usually improve the performance of
their original algorithms. The kernelized algorithms applying cross
validation obtain the best performance. They outperform the original
methods in 26 out of 27 datasets. The other two kernel versions of
the three original algorithms also have satisfiable performance. The
kernelized algorithms applying kernel alignment outperform the
original algorithms in 22 datasets and obtain an equal performance
in 3 datasets. Only 2 out of 27 datasets where the original
algorithms outperform the kernel algorithms applying kernel
alignment. Similarly, the kernelized algorithms applying the
unweighted kernel outperform the original algorithms in 18 datasets
and obtain an equal performance in 6 datasets. Only 3 out of 27
datasets where the original algorithms outperform the kernel
algorithms applying the unweighted kernel.

We note that although the cross validation method usually gives the
best performance, the other two kernel construction methods provide
comparable results in much shorter running time. For each dataset, a
run-time overhead of the kernelized algorithms applying cross
validation is of several hours (on Pentium IV 1.5GHz, Ram 1 GB)
while run-time overheads of the kernelized algorithms applying
aligned kernels and the unweighted kernel are about minutes and
seconds, respectively, for each dataset. Therefore, in time-limited
circumstance, it is attractive to apply an aligned kernel or an
unweighted kernel.

Note that the kernel alignment method are not appropriate for a
multi-modal dataset in which there may be several clusters of data
points for each class since, from eq.~\eqref{eq_align0}, the
function $\mbox{align}(K,Y)$ will attain the maximum value if and
only if all points of the same class are collapsed into a single
point. This may be one reason which explains why cross validated
kernels give better results than results of aligned kernels in our
experiments. Developing a new kernel alignment algorithm which
suitable for multi-modality is currently an open problem.

Comparing generalization performance induced by aligned kernels and
the unweighted kernel, algorithms applying aligned kernels perform
slightly better than algorithms applying the unweighted kernel. With
little overhead and satisfiable performance, however, the unweighted
kernel is still attractive for algorithms, like NCA (in contrast to
LMNN and DNE), which are not required a specification of target
neighbors $w_{ij}$. Since Euclidean distance with respect to the
unweighted kernel is usually not appropriate for specifying
$w_{ij}$, an KPCA-trick application of algorithms like LMNN and DNE
may still require some re-programming.

As noted in the previous section, aligned kernels usually does not
use all base kernels ($\alpha_i = 0$ for some $i$); in contrast, the
unweighted kernel uses all base kernels ($\alpha_i = 1$ for all
$i$). Hence, as described in Section~\ref{sect_unweighted}, the
feature space corresponding to the unweighted kernel usually
contains the feature space corresponding to aligned kernels.
Therefore, we may informally say that the feature space induced by
the unweighted kernel is ``larger'' than ones induced by aligned
kernels.

Since a feature space which is too large can lead to overfitting,
one may wonder whether or not using the unweighted kernel leads to
overfitting. Figure~\ref{fig_overfit} shows that overfitting indeed
does not occur. For compactness, we show only the results of
\textsc{Unweighted KDNE}. In the experiments shown in this figure,
base kernels are adding in the following order: 0.01, 0.025, 0.05,
0.075, 0.1, 0.25, 0.5, 0.75, 1, 2.5, 5, 7.5, 10, 25, 50, 75, 100,
250, 500, 750, 1000. It can be observed from the figure that the
generalization performance of \textsc{Unweighted KDNE} will be
eventually stable as we add more and more base kernels. Also, It can
be observed that 10 - 14 base kernels are enough to obtain stable
performance. It is interesting to further investigate an overfitting
behavior of a learner by applying methods such as a bias-variance
analysis \cite{James:mlj03} and investigate whether it is
appropriate or not to apply an ``adaptive resampling and combining''
method \cite{Breiman:AnStat98} to improve the classification
performance of a supervised mahalanobis distance learner.

\section{Summary}
\label{sect_summ} We have presented general frameworks to kernerlize
Mahalanobis distance learners. Three recent algorithms are
kernelized as examples. Although we have focused only on the
supervised settings, the frameworks are clearly applicable to
learners in other settings as well, e.g. a semi-supervised learner.
Two representer theorems which justify both our framework and those
in previous works are formally proven. The theorems can also be
applied to Mahalanobis distance learners in unsupervised and
semi-supervised settings. Moreover, we present two methods which can
be efficiently used for constructing a good kernel function from
training data. Although we have concentrated only on Mahalanobis
distance learners, our kernel construction methods can be indeed
applied to all kernel classifiers. Numerical results over various
real-world datasets showed consistent improvements of kernelized
learners over their original versions.

%In the future, we plan to extend the the current work to other
%settings such as semi-supervised, transductive and online settings.
%We also plan to investigate in details various types of regularizers
%appeared in the regularized versions of kernelized Mahalanobis
%distance learners proposed in Subsection~\ref{sect_rep}.

\subsubsection*{Acknowledgements} This work is supported by
Thailand Research Fund. We thank Wicharn Lewkeeratiyutkul who taught
us the theory of Hilbert space. We also thank Prasertsak
Pungprasertying for preparing some datasets in the experiments.

\bibliography{kmaha_arxiv09}
\bibliographystyle{theapa}
\end{document}